\documentclass{article}
\usepackage[preprint]{neurips_2026}

\usepackage[utf8]{inputenc} % allow utf-8 input
\usepackage[T1]{fontenc}    % use 8-bit T1 fonts
\usepackage{hyperref}       % hyperlinks
\usepackage{url}            % simple URL typesetting
\usepackage{booktabs}       % professional-quality tables
\usepackage{amsfonts}       % blackboard math symbols
\usepackage{nicefrac}       % compact symbols for 1/2, etc.
\usepackage{microtype}      % microtypography
\usepackage[table]{xcolor}

\usepackage{wrapfig}

\usepackage{graphicx}
\usepackage{booktabs}
\usepackage{subcaption}
\usepackage{wrapfig}
\usepackage{pifont}
\usepackage{arydshln}

\usepackage{hyperref}

\usepackage{orcidlink}
\usepackage{color,soul}
\usepackage{multirow}

\usepackage{amsmath} 

\usepackage[capitalize,noabbrev]{cleveref}
\usepackage{xspace}
\usepackage{algorithmic}
\usepackage{algorithm}

\makeatletter
\def\adl@drawiv#1#2#3{%
        \hskip.5\tabcolsep
        \xleaders#3{#2.5\@tempdimb #1{1}#2.5\@tempdimb}%
                #2\z@ plus1fil minus1fil\relax
        \hskip.5\tabcolsep}
\newcommand{\cdashlinelr}[1]{%
  \noalign{\vskip\aboverulesep
           \global\let\@dashdrawstore\adl@draw
           \global\let\adl@draw\adl@drawiv}
  \cdashline{#1}
  \noalign{\global\let\adl@draw\@dashdrawstore
           \vskip\belowrulesep}}
\makeatother

\title{Adaptive Subspace Projection for Generative Personalization}

\author{%
  Van-Anh Nguyen \\
  Monash University, Australia \\
  \texttt{van-anh.nguyen@monash.edu} \\
  \And
  Anh Bui \\
  Monash University, Australia \\
  \texttt{trunglm@monash.edu} \\
  \And
  Tamas Abraham \\
  Defence Science and Technology Group, Australia\\
  \texttt{tamas.abraham@defence.gov.au} \\
  \And
  Junae Kim \\
  Defence Science and Technology Group, Australia \\
  \texttt{junae.kim@defence.gov.au} \\
  \And
  Amardeep Kaur \\
  Defence Science and Technology Group, Australia \\
  \texttt{amar.kaur@defence.gov.au} \\
  \And
  Rollin Omari \\
  Defence Science and Technology Group, Australia \\
  \texttt{rollin.omari@defence.gov.au} \\
  \And
  Thuy-Trang Vu \\
  Monash University, Australia \\
  \texttt{Trang.Vu1@monash.edu} \\
  \And
  Dinh Phung \\
  Monash University, Australia \\
  \texttt{dinh.phung@monash.edu} \\
}

\begin{document}

\maketitle

\begin{abstract}
  Generative personalization often suffers from the semantic collapsing problem (SCP), where a learned personalized concept overpowers the rest of the text prompt, causing the model to ignore important contextual details. To address this, we first analyze the underlying cause, revealing that the semantic drift responsible for SCP is not random but is concentrated within a specific low-dimensional subspace. We also discover that the personalization process perturbs the embedding of the original base concept, making it an unstable reference point. Based on these insights, we introduce Test-time Embedding Adjustment with Adaptive Subspace Projection (AdaptSP), a training-free method that uses the stable, pre-trained embedding as an anchor. AdaptSP isolates the semantic drift and projects it onto the identified subspace, performing a precise adjustment that mitigates SCP while maintaining the subject identity. Our experiments show that this targeted approach significantly improves prompt fidelity and contextual alignment.
  
  % \keywords{Test-time alignment \and Generative personalization \and Semantic collapsing problem \and Subspace projection \and Residual}
\end{abstract}

\section{Introduction} \label{sec:intro}
Text-to-image (T2I) diffusion models, built on denoising diffusion and score-based formulations, now synthesize highly realistic images with strong text alignment at scale~\cite{ho2020denoising,song2020score,saharia2022imagen,rombach2022high}. These advances have fueled growing interest in generative personalization: adapting a pretrained T2I model to a user-specific concept (e.g., a person, pet, or object) from only a few reference images, while retaining the ability to place that concept into novel contexts via natural-language prompts~\cite{gal2022image,ruiz2023dreambooth}. The core objective is to preserve the unique identity of the personal concept while remaining faithful to the prompt’s semantics. Despite rapid progress, misalignment between the generated image and the input prompt remains a major challenge.

A robust generative personalization method should allow the user-defined visual concept to be composed with arbitrary contexts in the text prompt without losing fidelity or expressiveness. However, existing approaches often struggle to maintain prompt and generated image alignment, particularly with complex or multi-concept prompts~\cite{kong2024omg,zhu2025multibooth}. 
This misalignment has commonly been attributed to language drift or overfitting \cite{ruiz2023dreambooth}, arising from optimising on a limited number of reference images. 
Additional factors include limited expressiveness of textual embeddings \cite{zhang2023adding,mou2024t2i} and entangled reference sets \cite{avrahami2023break, jin2024image}, and semantic collapse \cite{bui2025mitigating}, where the learned personalized token loses its original textual semantics and over-dominates the prompt. Prior work mitigates these issues through regularization strategies \cite{qiu2023controlling,arar2024palp,motamed2024lego} or test-time embedding adjustment, which modifies the magnitude and direction of pretrained embeddings at inference to reduce semantic collapse \cite{bui2025mitigating}.
Compared to training-based regularization, test-time adjustment is lightweight, requires no retraining, and generalizes across personalization methods. However, it applies a global correction to the entire embedding and relies on post-personalization class embeddings as reference, which may introduce bias. It also overlooks structured, context-consistent residual drift, resulting in coarse adjustments for complex, multi-entity prompts.
To address these limitations, we propose a more targeted test-time adjustment method with 
\textbf{Adaptive Subspace Projection (AdaptSP)}.
% , guided by a simple decomposition: a personalized prompt embedding can be viewed as a class-based prompt embedding plus a concept-specific drift term that captures the identity of $V^*$. 
Our approach is motivated by a simple decomposition: a personalized prompt embedding can be viewed as a class-based prompt embedding plus a concept-specific drift term that captures the identity of the personalised concept. 
This drift can be extracted as the residual embedding between the personalized prompt and its original class-based prompt.
% To isolate this drift, we compare a personalized prompt with its class-swapped counterpart and define a residual
% \(r_i = \tau_{\phi}(\lfloor p_i, V^* \rfloor) - \tau_{\phi}(\lfloor p_i, c \rfloor)\).
Our preliminary analysis shows that these residuals are stable across diverse contexts and concentrate in a low-dimensional subspace.
% As observed in a comprehensive empirical study in Section \ref{subsec:residual_analysis}, these residuals are stable across diverse contexts and concentrate in a low-dimensional subspace. 
AdaptSP leverages this structure to decouple identity from context by conditioning on a class prompt while injecting only a projected residual.
%AdaptSP leverages this structure to decouple identity from context by conditioning on a class prompt while injecting only a projected residual, yielding
% \(\tau_{\phi_0}(\lfloor p_i, c \rfloor) + P_{\mathcal{R}}(r_i)\),
% where \(\tau_{\phi_0}\) is the original (pre-personalization) text encoder providing a clean anchor, and \(P_{\mathcal{R}}\) projects onto the concept-relevant residual subspace. 
By correcting only the structured drift component, AdaptSP preserves subject identity while restoring prompt fidelity in complex, multi-entity descriptions. 
Our proposed method is lightweight, training-free, and compatible with diverse personalization methods. Across two evaluated datasets, it consistently enhances text–image alignment without compromising subject fidelity in most settings.
% Empirically, on CC101 and CelebA we observe consistent gains in prompt alignment across TI/DB/CD baselines, with notable improvements such as +1.87 in $\text{CLIP}_{T}^{f}$ on CC101 for DB+AdaptSP, and +0.64 in $\text{CLIP}_{T}^{f}$ on CelebA for DB+AdaptSP (Table~\ref{tab:CC101_results} and Table~\ref{tab:celebA_results}). We also see improved visual fidelity in several settings (e.g., +4.6 CLIP-I for DB on CC101), with an expected trade-off in others (e.g., -2.37 CLIP-I on CelebA for DB+AdaptSP), while qualitative results confirm better prompt faithfulness (Section~\ref{sec:experiment}).
% Our approach is lightweight, training-free, and broadly compatible with different personalization methods, and it improves text-image alignment without sacrificing subject fidelity in most settings across CC101 and CelebA.

In summary, our contributions are as follows:
\begin{itemize}
    % \item We observe that the residual embeddings are remarkably consistent across diverse context prompts, indicating that they largely capture context-invariant information—primarily the subject identity encoded by $V^*$.
    \item We show that residual embeddings are highly consistent across diverse prompts, suggesting they primarily encode context-invariant subject identity.

    \item We represent the personalized concept as a residual embedding added to a class-based prompt embedding and introduce AdaptSP, a training-free, test-time method that decouples identity from context. This targeted adjustment mitigates prompt domination in the semantic collapsing problem and improves compositional fidelity in complex scenes.
    %This residual-based conditioning reduces the tendency of $V^*$ to dominate the overall prompt representation, thereby improving prompt fidelity and compositional generation in complex scenes.
    
    % \item We provide extensive evaluations on CC101 and CelebA with complex prompts, showing consistent gains in prompt alignment across standard generative personalization methods, with complementary ablations on subspace design and an analysis of subject-fidelity metric limitations.
    \item Through extensive experiments with complex prompts, we demonstrate consistent gains in prompt alignment across standard personalization methods, supported by ablation studies and analysis of subject-fidelity metrics.
\end{itemize}
% \textbf{\ding{182}} 
% \textbf{\ding{183}} 
% \textbf{\ding{184}} 

\section{Related Work} \label{sec:related_work}
\paragraph{Generative Personalization.}
% Generative personalization aims to bind a user-defined concept to a model so it can be rendered in new contexts. 
Generative personalization aims to bind a user-defined concept to a model for generation in new contexts. 
Classic subject-driven approaches include Textual Inversion (TI)~\cite{gal2022image} and DreamBooth~\cite{ruiz2023dreambooth}, and subsequent work explores parameter-efficient or targeted updates such as Custom Diffusion~\cite{kumari2023multi}, SVDiffusion~\cite{han2023svdiff}, OFT~\cite{qiu2023controlling}, and key-locking or rank-constrained tuning~\cite{tewel2023key}. Multi-concept and compositional personalization has been studied in methods like MultiBooth and related multi-subject settings~\cite{zhu2025multibooth,kong2024omg}.
Beyond subject-driven personalization, person/identity-driven methods focus on facial identity and composition, including FastComposer, Face0, DreamIdentity, and PhotoVerse~\cite{xiao2024fastcomposer,valevski2023face0,chen2024dreamidentity,chen2023photoverse}. Style-driven personalization explores controllable stylistic transfer, e.g., StyleDrop, StyleCrafter, and ArtAdapter~\cite{sohn2023styledrop,liu2023stylecrafter,chen2024artadapter}. Image- or adapter-based conditioning has also been explored to reduce tuning cost, such as IP-Adapter, PhotoMaker, and InstantID~\cite{ye2023ipadapter,li2023photomaker,wang2024instantid}, which inject image-derived identity cues through lightweight modules.

\paragraph{Challenges in Personalization.}
Personalization from a few examples faces several recurring challenges: language drift and overfitting, limited expressiveness of the conditioning signal, entanglement of concepts in reference sets, and prompt misalignment. Overfitting is commonly mitigated through prior preservation and regularization~\cite{ruiz2023dreambooth,han2023svdiff,qiu2023controlling,tewel2023key}. Limited expressiveness motivates adding new conditioning channels or adapters (e.g., ControlNet/T2I-Adapter/SCEdit)~\cite{zhang2023adding,mou2024t2i,jiang2024scedit}. Entanglement is addressed via masks, segmentation, or data augmentation~\cite{avrahami2023break,jin2024image,safaee2024clic,li2023generate,chen2023disenbooth}. Prompt misalignment and compositionality issues are studied in methods such as PALP, LEGO, and ReVersion~\cite{arar2024palp,motamed2024lego,huang2024reversion}.

\paragraph{Test-time Adjustment.}
% Bui \etal~\cite{bui2025mitigating} identify semantic collapse in personalized tokens and propose test-time embedding adjustment (TEA) to realign a learned token toward its anchor concept. 
Test-time adjustment aims to align both the magnitude and direction of personalized embeddings with their original semantic concepts during inference, thereby mitigating semantic collapse \cite{bui2025mitigating}. Prior work performs a global correction over the full prompt-embedding space using spherical linear interpolation (SLERP), interpolating between a personalized prompt embedding and its corresponding prior class embedding. This approach implicitly assumes that the undesired semantic drift is spread across the entire latent manifold and varies with each input prompt.
% In contrast, AdaptSP isolates the \emph{residual drift} between personalized and class-swapped prompts, shows this drift concentrates in a low-dimensional subspace, and performs a targeted projection within that subspace while anchoring context with the original (pre-personalization) text encoder. This yields a more precise, context-preserving adjustment that better balances subject fidelity with prompt fidelity in complex, multi-entity prompts.

\section{Semantic Collapsing in Personalized Generation} \label{sec:method}
% \subsection{Semantic collapsing problem} \label{subsec:smantic_colap}

% We construct a set of diverse contexts $\mathcal{P} = \{p_1, \dots, p_n\}$ to form two sets of prompts: the personalized set $\{ \lfloor p_i, V^* \rfloor \}_{i=1}^n$ and the anchor set $\{ \lfloor p_i, c \rfloor \}_{i=1}^n$. This gives us two corresponding sets of embeddings from the fine-tuned encoder: $\mathcal{S}_{V^*} = \{ \tau_{\phi}(\lfloor p_i, V^* \rfloor) \}_{i=1}^n$ and $\mathcal{S}_{c} = \{ \tau_{\phi}(\lfloor p_i, c \rfloor) \}_{i=1}^n$. We also define the residual (or drift) embedding set as $\mathcal{R} = \{ r_i:=  \tau_{\phi}(\lfloor p_i, V^* \rfloor) - \tau_{\phi}(\lfloor p_i, c \rfloor) \}_{i=1}^n$.

% - As mentioned in TEA, ... pc convert to v*, pv* convert to v*

% - It still happens even when the text encoder of c* left unchanged, with or without fine-tuning the unet. This is because text encoder of V* dominate the prompt embedding. For example, we learn a new context of a sks man generate image of sks man handshaking another man, this another man often have spacial charecteristic of the sks man, which resuce the diversity ...
% - V* also dominates the class c, which is related to V* and not included in prior_class or not change text embedding: V* is man, c is kid

% \subsection{Semantic Collapsing in Personalized Generation}
\label{sec:method_semantic_collapse}

\paragraph{Problem setup.}
We consider personalized text-to-image generation where a pre-trained diffusion model is adapted to represent a new subject using a learned concept token $V^*$ (e.g., \texttt{sks man}). 
% Let $c$ denote the corresponding anchor class token (e.g., \texttt{man}). \rebuttal{Given a prompt $p_c$ that contains $c$, and a corresponding prompt $p_{V^*}$ by replacing $c$ with $V^*$ while keeping the remaining text unchanged,} a text encoder $\tau_{\phi}$ maps any prompt into a sequence of token embeddings, which are then input to the U-Net for denoising.
Let $c$ denote the corresponding anchor class token (e.g., \texttt{man}). Given a class prompt $p_c$ that contains $c$ (e.g., \texttt{a man sitting on a red chair}), we construct the personalized prompt $p_{V^*}$ by replacing $c$ with the learned token $V^*$ while keeping the remaining text unchanged (e.g., a \texttt{sks} man sitting on a red chair). The text encoder $\tau_{\phi}$ maps each prompt into a sequence of token embeddings, which are then fed to the U-Net via cross-attention during the denoising process.

\paragraph{Semantic collapse in personalized tokens.}
Prior analyses \cite{bui2025mitigating} indicate that the personalized token can undergo semantic drift during personalization: $V^*$ gradually loses its original textual semantics while memorizing strong visual detail from the reference images. When $V^*$ appears in a prompt with rich descriptive content, the generated image can become disproportionately dominated by the personalized concept, neglecting other intended content and relations.
%
% Following this viewpoint, we characterize collapse through two prompt-to-embedding mappings:
% \begin{enumerate}
%     \item \textbf{Class-to-personalized conversion.} We compare the embeddings of the class prompt and the personalized prompt:
%     \begin{equation}
%         E(p_c) \;\rightarrow\; E(p_{V^*}),
%         \label{eq:pc_to_pv}
%     \end{equation}
%     which isolates the embedding shift introduced by substituting $c$ with $V^*$.
%     \item \textbf{Personalized self-consistency conversion.} We further transform the personalized prompt embedding toward a stabilized representation:
%     \begin{equation}
%         E(p_{V^*}) \;\rightarrow\; \widetilde{E}(p_{V^*}),
%         \label{eq:pv_to_pvtilde}
%     \end{equation}
%     where $\widetilde{E}(p_{V^*})$ is designed to preserve prompt semantics while retaining subject identity.
% \end{enumerate}
%
These conversions highlight that inserting $V^*$ may induce a global distortion of the prompt representation, rather than a localized replacement of a single token.

% \paragraph{Collapse persists with a fixed class embedding and across U-Net choices.}
A key observation is that semantic collapsing remains prevalent even when the embedding of the class token $c$ is left unchanged, as in several personalization pipelines such as Textual Inversion and Custom Diffusion variants, and regardless of whether the U-Net is fine-tuned. This suggests that collapse is not primarily caused by modifying the text encoder, nor does it require U-Net overfitting. Instead, the learned token $V^*$ can dominate the prompt embedding and the downstream cross-attention, effectively suppressing the influence of surrounding context tokens.

As an example, suppose we learn a concept token for \texttt{sks man}. For the prompt
\texttt{a sks man handshaking with a man},
the phrase \texttt{``a man''} should correspond to a different individual. In practice, the second person frequently inherits identity-like or spatial characteristics of the \texttt{sks man}  (e.g., similar facial structure or other subject-specific cues), despite being specified as a distinct entity. This leakage of subject traits into unrelated regions reduces diversity and corrupts multi-entity compositionality.

\begin{figure}[t]
\begin{subfigure}{1.0\textwidth}
  \centering
  \includegraphics[width=1\linewidth]{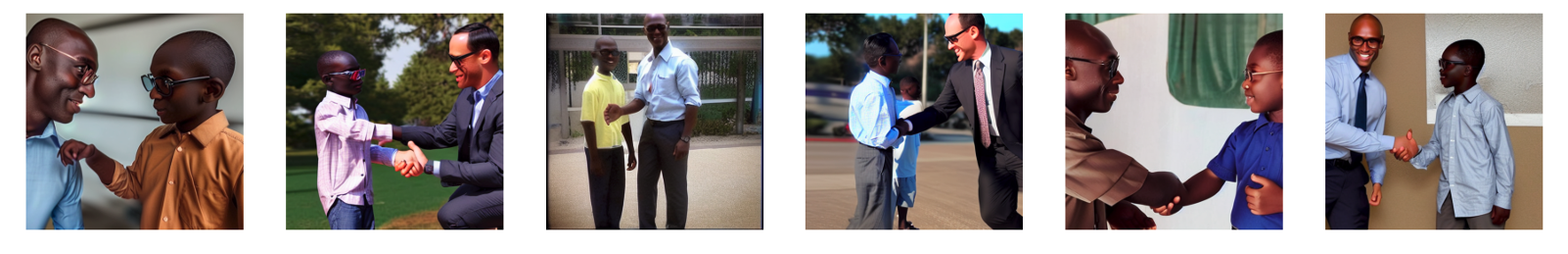}
    \caption{DreamBooth outputs}
    \label{fig:db_sample_kid}
\end{subfigure}%
\\
\begin{subfigure}{1.0\textwidth}
  \centering
  \includegraphics[width=1\linewidth]{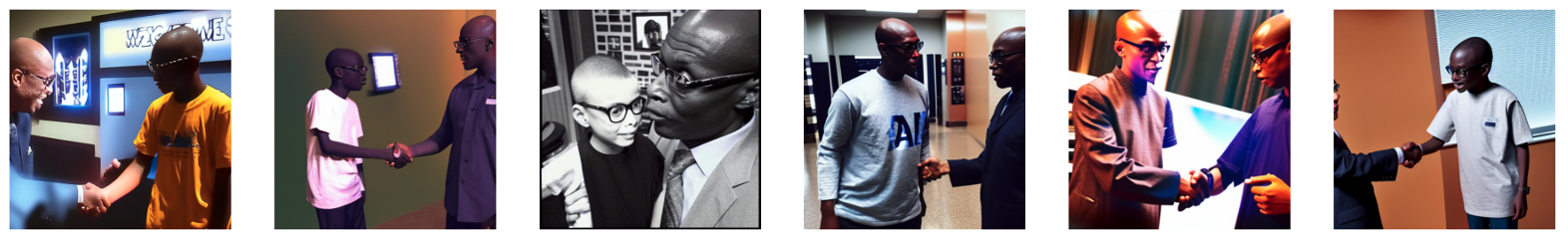}
    \caption{Custom-Diffusion outputs}
  \label{fig:cs_sample_kid}
\end{subfigure}
\caption{Example of SCP happening in either fine-tuning or without fine-tuning the embedding of existing classes, which affects not only the class $c$ ("man"), but also a related class "kid".}
\label{fig:baseline_sample}
\vspace{-1em}
\end{figure}

\paragraph{Cross-class dominance beyond the prior class set.}
We further find that the dominance of $V^*$ is not restricted to prompts containing the pre-defined anchor $c$. The personalized concept can also override related classes $c$ that are not part of the prior-class set and whose text embeddings are never updated. For example, when $V^*$ is learned for the class ``man'', prompts involving ``kid'' may still be pulled toward the subject's appearance distribution, producing a ``kid'' that partially adopts adult-like or subject-specific traits as presented in Figure \ref{fig:baseline_sample}. This cross-class interference indicates a broader imbalance in how personalized concepts are represented and attended to, beyond a narrow issue confined to the target class token.

% Give example of a man and kid
% Give examples of a capybara and a teddy bear

\paragraph{Method objective.}
Motivated by these findings, our methodology aims to control the dominance of $V^*$ in the prompt representation. The goal is to retain high subject fidelity when $V^*$ is used, while preventing the personalized token from overwhelming (i) contextual attributes and relations and (ii) other entities or classes mentioned in the prompt. In the next section, we introduce a mechanism that explicitly moderates the embedding $V^*$ to improve prompt fidelity.

\section{Analyzing the Residual Embedding and Subspace Structure}
\label{subsec:residual_analysis}

Although previous work ~\cite{bui2025mitigating} mitigates semantic collapse through test-time embedding adjustment, it simply assumes that semantic drift is evenly distributed across all dimensions and varies with each input prompt. However, high-dimensional embeddings often contain structured subspaces, where some directions encode subject identity while others capture context, relations, and compositional attributes \cite{zhao2026quantifying, bhalla2024interpreting}.

% Although previous work is effective at mitigating semantic collapsing~\cite{bui2025mitigating}, it relies on a single adjustment coefficient $\alpha$ applied uniformly across all embedding dimensions. This design implicitly assumes that all dimensions contribute equally to the undesired semantic drift. In practice, however, semantics in high-dimensional embedding spaces are often concentrated in structured subspaces: some directions encode the subject's identity (instance-specific visual cues), while others encode context, relations, and compositional attributes. A uniform adjustment can therefore be overly coarse, failing to isolate the dimensions that actually drive collapse. 

% To better localize the correction, we introduce an adaptive strategy that explicitly leverages this subspace structure. Our method, AdaptSP,
% % \textbf{Test-time Embedding Adjustment with Adaptive Subspace Projection (AdaptSP)},
% projects the adjustment onto a concept-relevant subspace, enabling targeted suppression of semantic drift while preserving context fidelity. We first analyze 

To better localize the correction and improve its efficiency, 
% we introduce AdaptSP, an adaptive strategy that explicitly leverages the structured subspace geometry of prompt embeddings. Instead of applying a uniform adjustment across the full embedding space, AdaptSP projects the correction onto a concept-relevant subspace, enabling targeted suppression of semantic drift while preserving contextual fidelity. W
we first analyze the residual drift induced by the personalized token and show that it is concentrated in a structured subspace, which motivates our adaptive subspace projection mechanism.

\subsection{Personalization induces a context-stable identity residual.}
\label{sec:method_rm}
Let $\tau_{\phi}$ denote the personalized text encoder, $V^*$ the learned concept token, and $c$ its corresponding class token. Given diverse context prompts $\{p_i\}_{i=1}^n$, we construct a personalized prompt $\lfloor p_i, V^* \rfloor$ and a class-swapped prompt $\lfloor p_i, c \rfloor$, where only $V^*$ is replaced by $c$. We define the residual embedding as
\begin{equation}
\label{eq:main_decompose_rewrite}
r_i
=
\tau_{\phi}(\lfloor p_i, V^* \rfloor)
-
\tau_{\phi}(\lfloor p_i, c \rfloor).
\end{equation}

The residual $r_i$ measures the embedding displacement introduced by the personalized token under the same context. Empirically, we find that these residuals are highly consistent across prompts, with average pairwise cosine similarity ranging from $0.3$ to $1.0$ and approximately $0.6$ on average. We can interpret each residual as $r_i = r_{\mathrm{id}} + \epsilon_i$,
where $r_{\mathrm{id}}$ denotes a context-invariant identity component associated with the personalized subject, and $\epsilon_i$ captures context-specific variation. The high similarity among residuals suggests that $r_{\mathrm{id}}$ dominates $\epsilon_i$, implying that the learned token $V^*$ contributes a stable identity direction or subspace across prompts.

This observation suggests that semantic collapse arises, in part, from a stable identity-related displacement that persists across contexts. We therefore approximate this component by the mean residual:
\begin{equation}
\label{eq:rm_def}
r_m
=
\frac{1}{n}\sum_{i=1}^n r_i.
\end{equation}

This gives the approximate decomposition:
\begin{equation}
\label{eq:second_decompose_rewrite}
\tau_{\phi}(\lfloor p_i, V^* \rfloor)
\approx
\tau_{\phi}(\lfloor p_i, c \rfloor) + r_m.
\end{equation}

This decomposition separates class-based contextual semantics from the identity information encoded by $V^*$. It suggests that personalization can be formulated as a controlled injection of an identity residual, rather than an entangled modification of the full prompt representation. Moreover, for each new subject, the mean residual $r_m$ only needs to be computed once from a small set of context prompts. At inference time, the same $r_m$ can then be reused across different prompts, preserving the efficiency of test-time adjustment while providing a more structured control over personalization-induced drift.

\paragraph{\textbf{Pre-personalization encoder as a stable semantic anchor.}}
Although Eq.~\eqref{eq:second_decompose_rewrite} separates the identity residual from the class-based prompt representation, $\tau_{\phi}(\lfloor p_i, c \rfloor)$ may still be biased by personalization, as discussed in Section~\ref{sec:method_semantic_collapse} and supported by the CEV analysis in Section~\ref{sec:method_pca}. To avoid relying on this potentially distorted anchor, we use the original text encoder $\tau_{\phi_0}$, which preserves the general language semantics before personalization. We therefore compute the class-prompt representation with $\tau_{\phi_0}$ and use it as a stable semantic anchor:

\begin{equation}
\label{eq:decompose_ori_text_en_rewrite}
\tau_{\phi}(\lfloor p_i, V^* \rfloor)
\approx
\tau_{\phi_0}(\lfloor p_i, c \rfloor) + r_m.
\end{equation}

\begin{figure}[h!]
    \centering
    \includegraphics[width=1\linewidth]{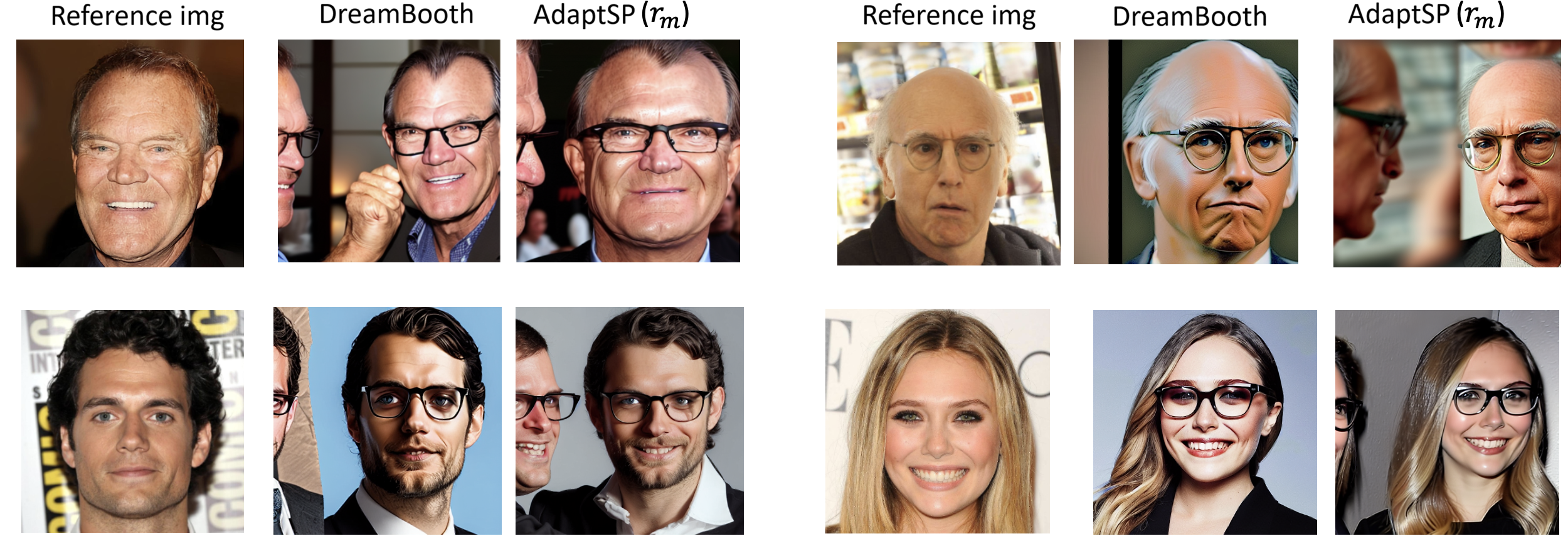}
    \caption{Comparing output of DreamBooth with and without AdaptSP ($r_m$) to demonstrate that the average of the residual embedding $r_m$ still contains salient features of concept $V^*$, which is prompt agnostic.}
    \label{fig:tea_pls_rm}
    % \vspace{-2em}
\end{figure}

% Here, $\tau_{\phi_0}(\lfloor p_i, c \rfloor)$ preserves the original compositional semantics, while $r_m$ injects the personalized identity. 
% This formulation enables AdaptSP to target personalization-induced drift without applying a global correction over the full embedding space.
We further validate this decomposition in Figure~\ref{fig:tea_pls_rm}. The reconstructed embeddings preserve the personalized subject while better maintaining contextual attributes and secondary entities than direct conditioning on $\lfloor p_i, V^* \rfloor$.

\subsection{Residual Drift Lies in a Low-Dimensional Subspace}
\label{sec:method_pca}
The strong clustering of residual embeddings suggests that personalization-induced drift is not arbitrary in the full embedding space, but concentrated along a few dominant directions. This motivates a subspace-based formulation: instead of representing the personalized concept only by the mean residual $r_m$, we model residual variation within a concept-relevant subspace. This allows AdaptSP to retain subject-specific information while preserving prompt-dependent semantics outside the dominant drift subspace.

\begin{figure}
    \begin{subfigure}[b]{0.49\textwidth}
    \centering
    \includegraphics[width=1.0\textwidth]{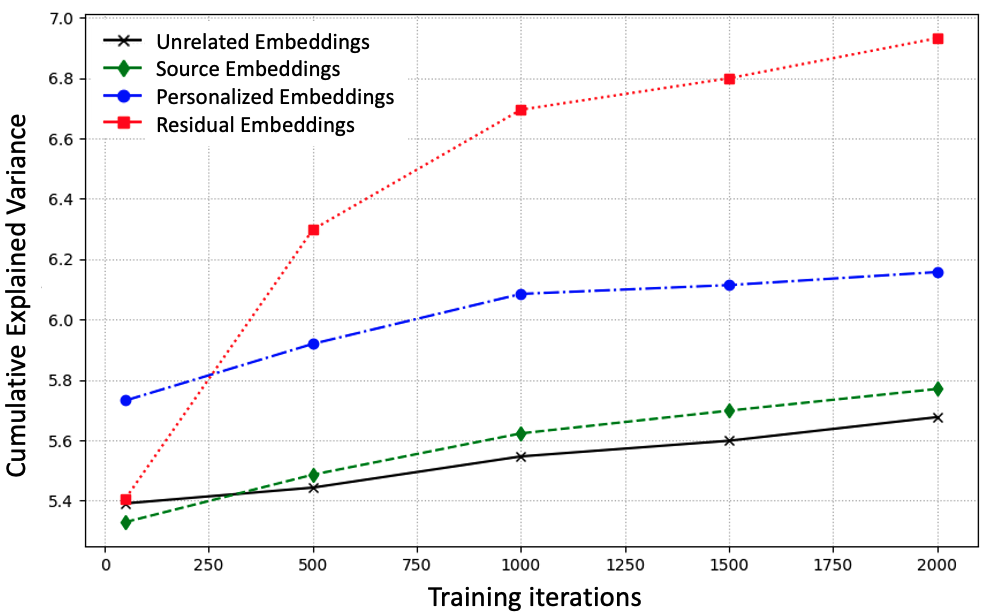}
        \caption{}
        \label{fig:pca_cumulative_explained_variance_celebA}
    \end{subfigure}
    \begin{subfigure}[b]{0.49\textwidth}
        \centering
        \includegraphics[width=1.0\textwidth]{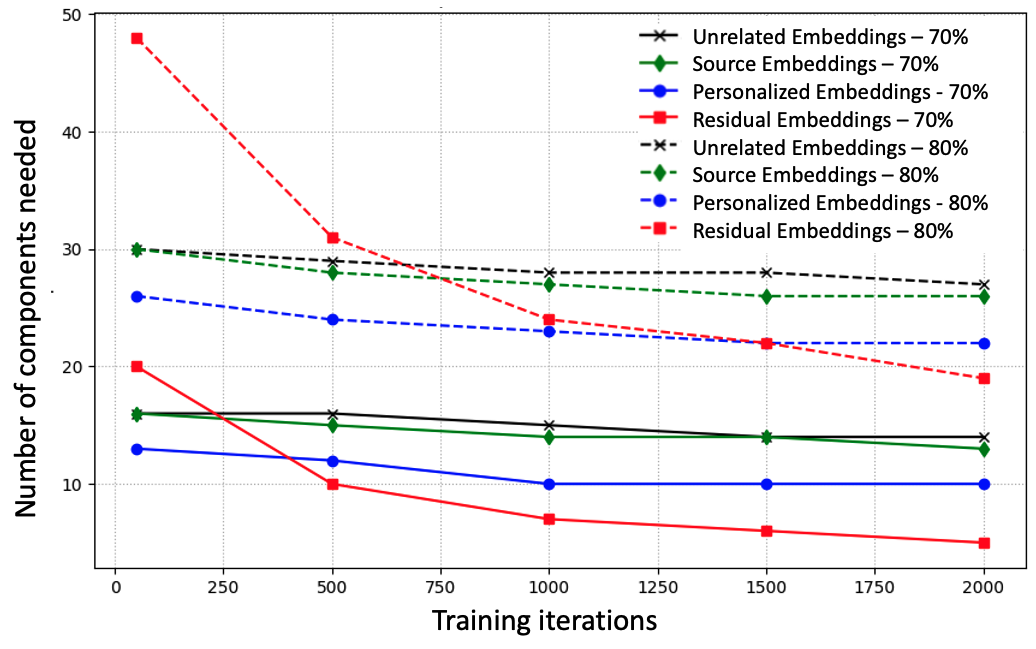}
        \caption{}
        \label{fig:pca_number_of_components_needed_celebA}
    \end{subfigure}
    \caption{Analysis on the CelebA dataset (a) Cumulative explained variance by the first 10 principal components. (b) Number of components needed to exceed the cumulative explained variance threshold of 70\% and 80\%.}
    \label{fig:pca_embeddings_celebA}
    % \vspace{-1em}
\end{figure}

\begin{figure}
    \begin{subfigure}[b]{0.49\textwidth}
    \centering
    \includegraphics[width=1.0\textwidth]{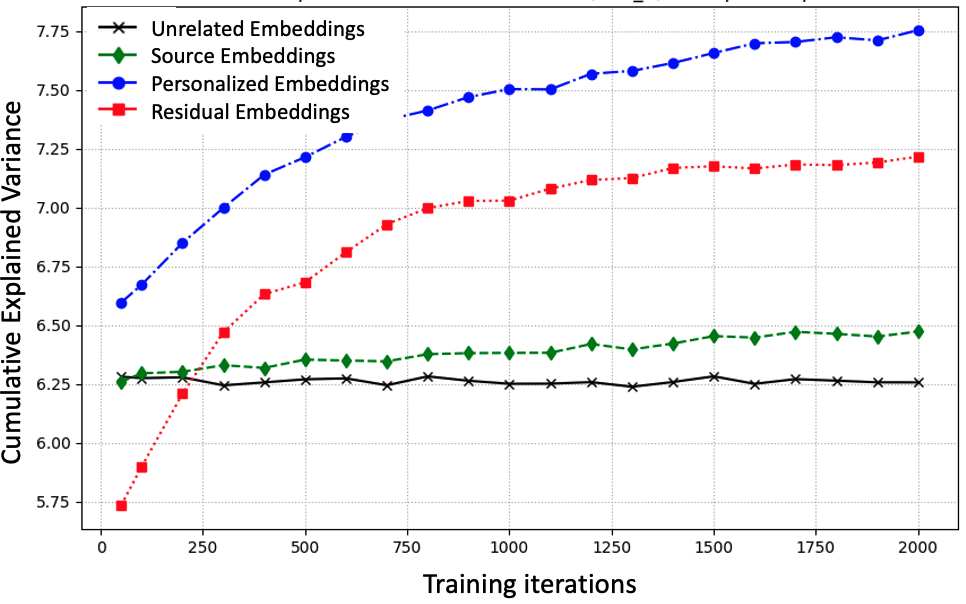}
        \caption{}
        \label{fig:pca_cumulative_explained_variance_CC101}
    \end{subfigure}
    \begin{subfigure}[b]{0.49\textwidth}
        \centering
        \includegraphics[width=1.0\textwidth]{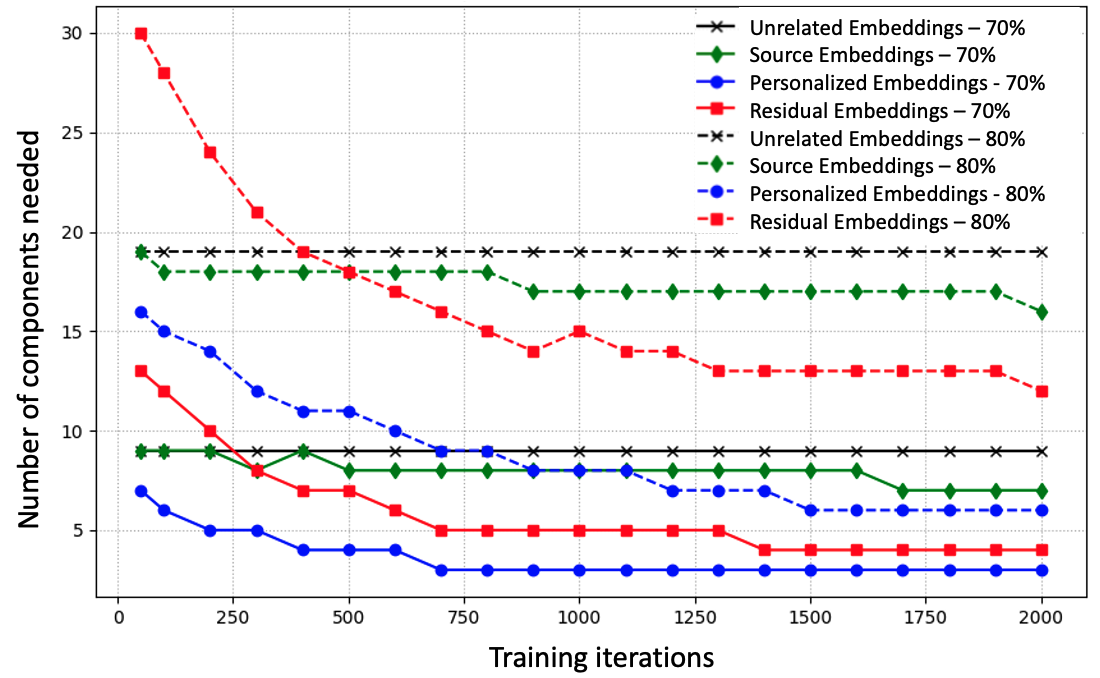}
        \caption{}
        \label{fig:pca_number_of_components_needed_CC101}
    \end{subfigure}
    \caption{Analysis in the CC101 dataset (a) Cumulative explained variance by the first 10 principal components. (b) Number of components needed to exceed the cumulative explained variance threshold of 70\% and 80\%.}
    \label{fig:pca_embeddings_CC101}
\end{figure}

To characterize this structure, we perform Principal Component Analysis (PCA) and compare the cumulative explained variance (CEV) of three embedding sets:
\begin{align}
\text{Source embeddings:} \quad 
& \mathcal{S}_{c} := \{\, \tau_{\phi}(\lfloor p_i, c \rfloor)\, \}_{i=1}^n, \\
\text{Personalized embeddings:} \quad  
& \mathcal{S}_{V^*} := \{\, \tau_{\phi}(\lfloor p_i, V^* \rfloor)\, \}_{i=1}^n, \\
\text{Residual embeddings:} \quad  
& \mathcal{R} := \{\, r_i = \tau_{\phi}(\lfloor p_i, V^* \rfloor) - \tau_{\phi}(\lfloor p_i, c \rfloor)\, \}_{i=1}^n .
\end{align}
Figures~\ref{fig:pca_embeddings_celebA} and~\ref{fig:pca_cumulative_explained_variance_CC101} show the CEV curves on CelebA and CC101. A faster-rising CEV curve indicates that the variation is captured by fewer principal directions.

We observe three consistent patterns. First, $\mathcal{S}_{V^*}$ exhibits higher concentration after personalization, suggesting reduced contextual diversity in the prompt representation. Second, $\mathcal{R}$ shows the strongest concentration, indicating that the transformation from $c$ to $V^*$ is largely captured by a low-dimensional residual subspace. Third, $\mathcal{S}_{c}$ also becomes more concentrated, suggesting that personalization can indirectly affect nearby class representations under the fine-tuned encoder $\tau_{\phi}$, even when $c$ is not explicitly optimized.

These findings motivate modeling personalization-induced drift using a residual subspace. While $r_m$ captures the shared identity component, residual variations across prompts contain additional structure related to context-dependent expressions of the personalized concept. The leading principal components of $\mathcal{R}$ capture these dominant directions, providing a compact and concept-relevant basis for prompt-specific correction. We therefore construct a subspace from these components and project only the residual component:

\begin{equation}
\label{eq:residual_projection_form}
\tau_{\phi}(\lfloor p_i, V^* \rfloor)
\approx
\tau_{\phi}(\lfloor p_i, c \rfloor)
+
P_{\mathcal{R}}(r_i),
\end{equation}
where $P_{\mathcal{R}}(\cdot)$ denotes projection onto the dominant residual subspace constructed by PCA. By controlling the number of retained principal components, we can adjust the capacity of this subspace: a smaller subspace enforces stronger suppression of personalization-induced drift, while a larger subspace preserves more concept-specific variation. This provides a flexible mechanism for balancing subject fidelity and prompt consistency. This residual-subspace formulation is a key component of AdaptSP.

\paragraph{Connection to anchor stability.}
The mild increase in the CEV of $\mathcal{S}_c$ indicates that the class-prompt embedding $\tau_{\phi}(\lfloor p, c \rfloor)$ may also drift under personalization, making it an imperfect reference, as discussed in Section~\ref{sec:method_rm}. Following Eq.~\eqref{eq:decompose_ori_text_en_rewrite}, we replace this potentially biased anchor with the class-prompt embedding computed by the original text encoder $\tau_{\phi_0}$, leading to the AdaptSP($P_{\mathcal{R}}$) formulation:

\begin{equation}
\label{eq:residual_projection_form_origin_tau}
\tau_{\phi}(\lfloor p_i, V^* \rfloor)
\approx
\tau_{\phi_0}(\lfloor p_i, c \rfloor)
+
P_{\mathcal{R}}\!
(r_i),
\end{equation}

%Note: can change that using $\tau_{\phi_0}$ for DreamBooth only because only Dreambooth change text encoder (changes all embedding of all words)

\section{Experiments} \label{sec:experiment}
\subsection{Experimental Setup}

\textbf{Reference Images.} 
We use a subset of 9 concepts from the CustomConcept101 (CC101) dataset as in the original paper \cite{kumari2023multi}, 
each of which has 3-15 images, including  `Barn', `Tortoise plushy', `Teddy-Bear', `Wooden Pot', `Dog', `Cat', `Flower', `Table', `Chair' subjects.
For the human concept, we use a subset of 10 concepts from the CelebA dataset \cite{liu2015faceattributes}, which includes 10 identities with 10-15 images per subject (different women and men).
% Sample images from the two datasets are shown in Figures~\ref{fig:sample_CC101_dataset} and~\ref{fig:sample_celeba_dataset}.

\noindent\textbf{Prompts.}
% We collect complex prompts from the CC101 dataset, where each prompt contains at least two concepts, e.g.,  
% % `a \textcolor{blue}{watercolor painting} of \textcolor{red}{$V^*$ tortoise plushy} \textcolor{blue}{on a mountain}'. 
% \texttt{a photo of {$V^*$ tortoise plushy} {on a mountain}}. 
% For the human concept, we create a list of 17 prompts, where each prompt contains the main concept and a complex context/action, 
% e.g., \texttt{A photo of a {$V^*$} wearing glasses and {writing in a red notebook}}. 
We use complex prompts from the CC101 dataset, where each prompt contains at least two concepts. For human subjects, we additionally construct a set of 17 prompts, each combining the personalized subject with a complex context or action. These prompts are designed to evaluate both subject identity preservation and prompt fidelity.

To construct the identity subspace $P_{\mathcal{R}}$ and the subject-specific residual representation $r_m$, we generate a small auxiliary set of 60-100 prompts for each dataset, CC101 and CelebA, using an LLM~\cite{OpenAI_ChatGPT_2026}. These prompts are not tailored to individual subjects but instead provide general contextual variations for estimating residual structure efficiently. We provide some examples in Appendix \ref{tab:sample_prompts_subspace}.

\noindent\textbf{Pre-trained T2I models.}
We conduct experiments using two pre-trained text-to-image backbones: Stable Diffusion v1.5 and FLUX.1-dev. The specific backbone used in each experiment is specified in the corresponding experimental setting.

\noindent\textbf{Metrics.} We evaluate personalization quality using CLIP-based text-image and image-image alignment scores. For prompt fidelity, we report CLIP-T in two settings: CLIP-T$^{f}$, computed with the full prompt $\lfloor p, c \rfloor$, and CLIP-T$^{p}$, computed with the context prompt $p$ only. For subject fidelity, we report CLIP-I, which measures similarity between generated images and reference subject images using CLIP visual features~\cite{radford2021learning,hessel2021clipscore}. We also report DINO image-image similarity in a self-supervised feature space~\cite{caron2021emerging}.

Although these metrics are commonly used for personalized generation, they have known limitations. We discuss these limitations in Section~\ref{subsec:metric} and provide human evaluation to better assess perceptual quality, subject preservation, and prompt consistency.

\subsection{Evaluation Results}

We conducted extensive experiments to validate the effectiveness of our proposed method, AdaptSP, when combined with representative personalization baselines, including DreamBooth~\cite{ruiz2023dreambooth}, Custom Diffusion~\cite{kumari2023multi}, and Class Diffusion~\cite{huang2025classdiffusion}.

\begin{table}[h!]
\caption{Comparison of average performance on CelebA and CC101 when integrating AdaptSP with personalization baselines. We use Stable-Diffusion-v1-5 as a pre-trained text-to-image model for all experiments.
    }
    \label{tab:all_result}
    \centering
    \resizebox{1.0\columnwidth}{!}{
\begin{tabular}{cccccccccc}
\toprule
           & \multicolumn{4}{c}{\multirow{2}{*}{\textbf{CelebA dataset}} }                             &  & \multicolumn{4}{c}{\multirow{2}{*}{\textbf{CC101 dataset}}}                              \\ 
           & \\
           & \quad Baseline        & \quad TEA & \quad AdaptSP ($r_m$)  & \quad AdaptSP ($P_{\mathcal{R}}$) & \quad \quad  & Baseline       & \quad TEA & \quad AdaptSP ($r_m$) & \quad AdaptSP ($P_{\mathcal{R}}$) \\ \midrule \midrule
\multicolumn{10}{l}{\multirow{2}{*}{\textbf{DreamBooth}}}               \\                                                                                                    \\
Clip-T$^p$ & 19.68         & 20.19   & \textbf{20.93} & 20.44                      &  & 14.49      &      15.26      & \textbf{18.50}   & 17.31                       \\
Clip-T$^f$ & 24.56      &  25.20    & \textbf{26.35} & 25.44                      &  & 21.93    &       23.80       & \textbf{25.76}  & 24.47                       \\ \cdashlinelr{1-10}
\rowcolor{lightgray!50} Clip-I     & 61.98     &    59.61   & 61.16          & \textbf{63.40}             &  & 85.45 &   \textbf{89.45}     & 78.33           & 79.63                       \\
\rowcolor{lightgray!50} DINO       & \textbf{47.34} &  45.07 & 44.30          & 46.89                      &  & 71.16 &     \textbf{71.75}   & 59.68           & 62.09                       \\ \midrule
\multicolumn{10}{l}{\multirow{2}{*}{\textbf{Class Diffusion}}}    \\                                                                                                          \\
Clip-T$^p$ & 21.28     &   21.48    & \textbf{21.50} & 19.70                      &  & 21.41     &     21.29        & \textbf{22.15}  & 20.76                       \\
Clip-T$^f$ & 27.41      &  27.80    & \textbf{27.91} & 25.22                      &  & \textbf{26.23} &    26.65    & 25.98           & 25.38                       \\ \cdashlinelr{1-10}
\rowcolor{lightgray!50} Clip-I     & \textbf{57.33} &    56.64   &  55.22          & 55.62                      &  & 66.74       &    67.04    & 62.72           & \textbf{67.10}               \\
\rowcolor{lightgray!50} DINO       & 39.44      &    \textbf{39.60}   & 37.45          & 39.62             &  & 42.00   &     \textbf{43.24}    & 30.91           & 42.04                       \\ \midrule
\multicolumn{10}{l}{\multirow{2}{*}{\textbf{Custom Diffusion}}}    \\                                                                                                         \\
Clip-T$^p$ & 21.45         &  21.33  & \textbf{21.85}   & 21.15                       &  & 20.53      &      20.41      & \textbf{21.44}  & 19.36                       \\
Clip-T$^f$ & 26.96       &   27.05   & \textbf{28.40} & 27.74                      &  & 26.88    &       27.22       & 26.93  & 25.54                       \\  \cdashlinelr{1-10}
\rowcolor{lightgray!50} Clip-I & \textbf{56.27} &  55.58    &      53.09          & 55.70                      &   & 72.20             &    72.57       & 67.78           & \textbf{72.65}              \\
\rowcolor{lightgray!50} DINO    & 37.66   &   \textbf{40.75}    & 33.18           & 35.87        &        & 47.81          &      48.05     & 38.51           & \textbf{49.42}         \\ 
\bottomrule
\end{tabular}
}
\end{table}

For AdaptSP($P_{\mathcal R}$), we retain the top 2 principal components on CelebA and the top 5 on CC101. Table~\ref{tab:all_result} reports the average results on both datasets after integrating AdaptSP with different baselines; all experiments use the same Stable-Diffusion-v1-5 as a pre-trained text-to-image model. Detailed CelebA results are shown in Table~\ref{tab:celebA_results}, with the full CC101 breakdown provided in the supplementary material. Overall, AdaptSP consistently improves prompt fidelity in most settings, supporting our claim that residual-based conditioning reduces personalized-concept overemphasis and better preserves contextual tokens. Additional experiments with the pre-trained FLUX.1-dev model is presented in Appendix \ref{sec:appendix_experiments}.

The two AdaptSP variants offer complementary trade-offs. AdaptSP($r_m$) generally achieves stronger prompt-alignment gains with minimal additional computation, preserving inference efficiency. AdaptSP($P_{\mathcal R}$) better preserves identity in some cases. On CC101, AdaptSP($P_{\mathcal R}$) also improves subject-fidelity metrics for Class Diffusion and Custom Diffusion, suggesting that residual subspace projection can retain identity cues while improving prompt consistency. 

We note that image-image metrics on CC101 can be misleading for DreamBooth due to overfitting. The model may reproduce training-specific backgrounds, colors, or textures, inflating CLIP-I/DINO despite weak compositional generalization. In such cases, AdaptSP may reduce CLIP-I/DINO while improving prompt fidelity, as generations become more diverse and less tied to the training scenes. We discuss this further in Section~\ref{subsec:metric}.
\vspace{-1em}

\begin{table}[h!]
\centering
    \caption{Detail results on the CelebA dataset. The first/\colorbox{lightgray!50} {second} metric is the $\text{CLIP-T}^{f}$/\colorbox{lightgray!50}{CLIP-I} score. The highest score is shown in \textbf{bold}, and the runner-up is \underline{underlined}.
    % The \textcolor{blue}{blue number} indicates the proposed method outperforms its baseline counterpart, while the \textcolor{red}{red number} indicates the opposite. 
    % The GAP is the average improvement over all concepts. Qualitative results are shown in Fig. \ref{fig:qualitative_comparison_celeba}.
    }
    \label{tab:celebA_results}
    % \centering
    \resizebox{1.0\textwidth}{!}{
    \begin{tabular}{lccccccccccc}
    \toprule
    Method & 124 & 181 & 276 & 342 & 351 & 437 & 615 & 908 & 1335 & 1429 & Average\\
    \midrule
    % Textual Inversion  & 23.70    &  26.29    &  21.32    &  26.06    &  25.25    &  23.46    &  26.54    &  24.61   &  25.40    &  24.97  \\
    %     & \rowcolor{lightgray!50}  68.93  &  61.68  &  65.33  &  75.64  &  69.63  &  71.19  &  74.37  &  64.50 &  67.55  &  60.88  \\
    % Textual Inversion+$TEA$  & 24.23    &  26.49   & 20.97   &  27.28    &  26.11    &  23.93    &  26.89    &  25.98    &  25.69    &  25.68 \\
    %          & \rowcolor{lightgray!50}  68.48  &  59.60  &  65.20   &  71.01  &  65.73  &  66.25  &  70.56  &  62.97  &  66.72  &  59.01  \\
    % \midrule
    DreamBooth  & 25.69   &    25.42   &    22.40   &    26.55   &    26.54   &    20.40   &    26.64   &    22.73   &    23.94   &    25.33   &    24.56    \\
     \rowcolor{lightgray!50} \cellcolor{white} &  62.84  & 61.58  &  58.33  &  62.81  &  59.45  &  65.52  &  58.91  &  66.87  &  64.27  &  59.29 & \underline{61.98} \\
    DreamBooth+$TEA$  & 26.26    &  26.59    &  24.30    &  25.58    &  25.43    &  22.56    &  26.03    &  25.92    &  24.79    &  25.37  & 25.28 \\
            \rowcolor{lightgray!50} \cellcolor{white} &  50.67  &  60.23  &  47.66  &  65.83  &  61.34  &  68.79  &  60.88  &  60.34  &  53.66  &  58.46 & 58.79 \\
             \cdashlinelr{1-12}

    DreamBooth+$AdaptSP$ \\
    \quad \quad ($r_m$)  &   26.22   &    26.99   &    24.44   &    27.51   &    27.74   &    23.73   &    27.55   &    26.34   &    26.53   &    26.54   &    \textbf{26.36}     \\
\rowcolor{lightgray!50} \cellcolor{white} & 62.50  &  60.50  &  62.05  &  63.42  &  58.87  &  65.89  &  58.89  &  63.91  &  57.91  &  57.69  &  61.16 \\
    \quad \quad ($P_{\mathcal{R}}$)  &   25.74   &    26.35   &    22.55   &    26.43   &    27.19   &    22.46   &    26.95   &    24.70   &    25.84   &    26.25   &    \underline{25.45}     \\
   \rowcolor{lightgray!50} \cellcolor{white} & 63.52  &  62.07  &  65.98  &  69.14  &  60.44  &  68.96  &  59.01  &  66.65  &  60.13  &  58.11  &  \textbf{63.40} \\ \midrule
    Class Diffusion  & 26.88 & 28.36 & 26.93 & 27.26 & 28.59 & 24.9 & 27.61 & 27.64 & 27.27 & 28.67 & 27.41    \\
\rowcolor{lightgray!50} \cellcolor{white} & 60.95 & 49.51 & 53.77 & 56.62 & 57.43 & 67.84 & 56.48 & 58.13 & 56.3 & 56.29 & \textbf{57.33}\\

% Class Diffusion+$TEA$  & 27.09    &  26.83    &  27.18    &  27.37    &  26.99    &  27.11    &  27.16    &  26.93    &  26.94    &  26.98 & 27.06   \\
%         &  \rowcolor{lightgray!50} 44.88  &  55.03  &  37.10  &  53.63  &  55.56  &  56.41  &  55.01  &  55.33  &  46.68  &  52.27 & 51.19\\ 
\cdashlinelr{1-12}
    Class Diffusion+$AdaptSP$ \\
    \quad \quad ($r_m$) & 27.7 & 28.81 & 27.39 & 27.61 & 28.58 & 25.53 & 27.81 & 27.98 & 28.12 & 29.63 & \textbf{27.92} \\
\rowcolor{lightgray!50} \cellcolor{white} & 59.49 & 49.66 & 52.74 & 56.19 & 53.56 & 62.76 & 53.82 & 55.52 & 53.96 & 54.52 & 55.22  \\

    \quad \quad ($P_{\mathcal{R}}$) & 27.50  &  28.69  &  27.45  &  27.41  &  28.43  &  26.33  &  27.66  &  27.64  &  27.54  &  28.77  &  \underline{27.74} \\
 \rowcolor{lightgray!50} \cellcolor{white} & 58.74  &  50.94  &  53.07  &  55.10  &  54.32  &  61.51  &  55.76  &  56.04  &  55.44  &  56.10  &  \underline{55.70}  \\ \midrule
    Custom Diffusion  & 26.92    &  26.87    &  26.91    &  27.14    &  26.94    &  26.97    &  26.99    &  26.93    &  27.05    &  26.88 & 26.96    \\
% \rowcolor{lightgray!50} \cellcolor{white} & 43.56  &  52.48  &  36.22  &  52.64  &  53.39  &  55.45  &  53.96  &  53.32  &  44.95  &        50.33  & 49.63\\

 \rowcolor{lightgray!50} \cellcolor{white} & 60.31 & 50.33 & 52.88 & 56.03 & 56.38 & 62.34 & 56.83 & 56.82 & 54.58 & 56.25 & \textbf{56.27} \\

Custom Diffusion+$TEA$  & 27.09    &  26.83    &  27.18    &  27.37    &  26.99    &  27.11    &  27.16    &  26.93    &  26.94    &  26.98 & 27.06   \\
  \rowcolor{lightgray!50} \cellcolor{white}        & 44.88  &  55.03  &  37.10  &  53.63  &  55.56  &  56.41  &  55.01  &  55.33  &  46.68  &  52.27 & 51.19 \\ \cdashlinelr{1-12}
    Custom Diffusion+$AdaptSP$ \\
    \quad \quad ($r_m$) & 28.23  &  29.10  &  27.86  &  27.91  &  28.60  &  27.58  &  27.87  &  28.22  &  28.94  &  29.71  &  \textbf{28.40} \\
 \rowcolor{lightgray!50} \cellcolor{white}     &   54.65  &  49.94  &  48.83  &  53.64  &  53.50  &  54.88  &  54.75  &  54.53    &  52.15  &  54.07  &  53.10  \\

    \quad \quad ($P_{\mathcal{R}}$) & 27.50  &  28.69  &  27.45  &  27.41  &  28.43  &  26.33  &  27.66  &  27.64  &  27.54  &  28.77  &  \underline{27.74} \\
 \rowcolor{lightgray!50} \cellcolor{white}     &  58.74  &  50.94  &  53.07  &  55.10  &  54.32  &  61.51  &  55.76  &  56.04  &  55.44  &  56.10  &  \underline{55.70}  \\
    \bottomrule
    \end{tabular}
    }
    % \vspace{-1em}
\end{table}

% Moreover, AdaptSP($P_{\mathcal{R}}$) further strengthens subject fidelity by projecting the individual residual embedding onto the learned subspace. On DreamBooth, AdaptSP($P_{\mathcal{R}}$) improves CLIP-I from $61.98$ to 63.40 while still improving $\text{CLIP-T}^{f}$; on Custom Diffusion, it yields a large gain in CLIP-I from $50.33$ to 55.70. These results validate our second observation that semantic drift concentrates in a low-dimensional subspace, and that subspace projection provides a more targeted correction than using $r_m$ alone.

% Experiments on the CustomConcept101 (CC101) dataset consistently improve prompt fidelity when applied to DreamBooth and CD. This supports our key claim that a residual-based representation can retain the core semantic/identity features of the personalized concept $V^*$, while reducing the dominance of $V^*$ in the prompt embedding and thereby improving adherence to context.

% We also observe that CLIP-I on CC101 can be less reliable, especially for DreamBooth. Since DB can overfit on CC101, it often reproduces training-specific cues (e.g., similar background, color tone, and textures), which can inflate image-image similarity scores. By encouraging more context-faithful and diverse generations, AdaptSP may reduce these spurious similarities, leading to a lower CLIP-I despite improved qualitative results. We further discuss this evaluation issue in Section~\ref{subsec:metric}.

\newcommand{\win}[1]{\textcolor{black}{#1}}
\newcommand{\lose}[1]{\textcolor{black}{#1}}
\newcommand{\abnormal}[1]{\textcolor{red}{#1}}

\section{Ablation Study}

\subsection{Impact of number of principal components}

\begin{wrapfigure}{r}{0.5\textwidth}
    \centering
    \includegraphics[width=1\linewidth]{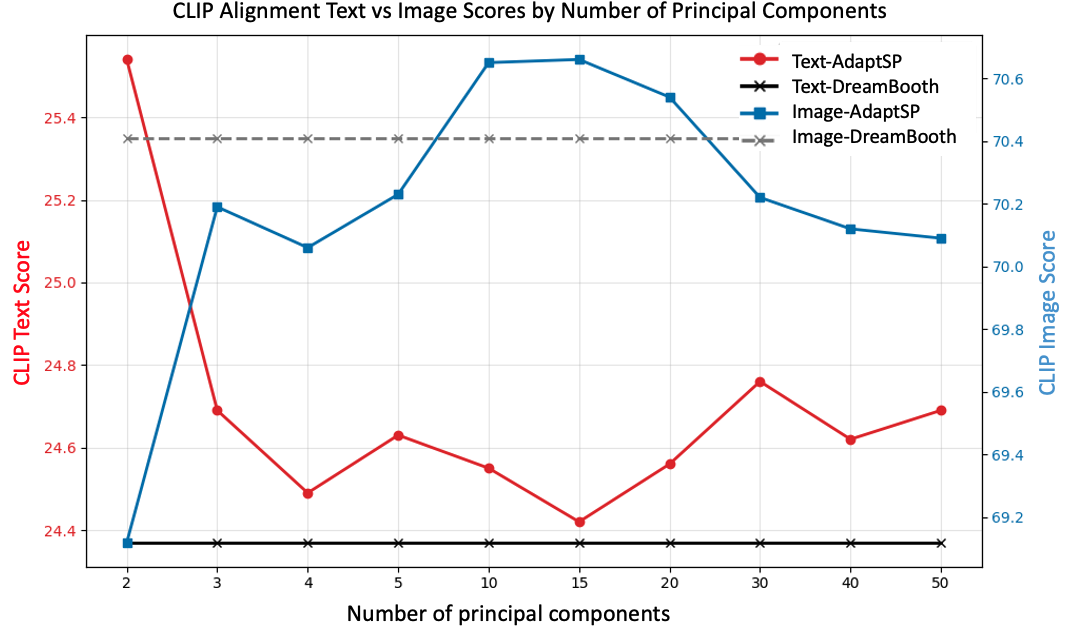}
    \caption{Impact of the number of principal components (PCs) on scores}
    \label{fig:ablation_study_number_of_principal_components}
    \vspace{-1em}
\end{wrapfigure}

Figure~\ref{fig:ablation_study_number_of_principal_components} studies the effect of the number of principal components (PCs) used to define the residual subspace. As the PC count increases, text alignment decreases initially and then stabilizes, while image alignment improves rapidly and even passes the baseline.

This suggests a trade-off controlled by the subspace dimension: more PCs preserve stronger identity information, but too many can reintroduce excessive concept drift and weaken prompt alignment. Empirical experiments show that using a small number of PCs, around 2-5, provides the best balance, which retains useful identity cues while filtering less relevant or collapse-inducing directions.

% From the above analysis, we can conclude that when: 
% \begin{itemize}
%     \item Preferring Textual Followability: Using a very small number of PCs (less than 5), the textual followability is significantly improved, but at the cost of the image alignment score.
%     \item Preferring Personalized Visual Followability: Due to the noisy nature of the embedding space, there exists an optimal number of PCs that can achieve a better image alignment score than the original embedding space.
% \end{itemize}

\subsection{Impact of using original pre-trained text encoder}
% \subsection{Impact of Using a Stable Pre-trained Anchor Encoder}
\label{subsec:anchor_encoder}

% Recall from Section \ref{residual_analysis} that our residual-based formulation decomposes a personalized prompt embedding into (i) an identity drift component and (ii) an anchor embedding that should represent the general context and compositional structure of the prompt. A natural option is to compute the anchor with the post-personalization text encoder $\tau_{\phi}$, i.e., $\tau_{\phi}(\lfloor p, c \rfloor)$. However, we find that this choice can be unreliable: the personalization process that learns $V^*$ may also distort the embedding space around the class concept, even when $c$ itself is not explicitly optimized.

Recall from Section~\ref{subsec:residual_analysis} that our personalized prompt embedding formulation is represented as the sum of (i) an identity drift component and (ii) an anchor embedding that should preserve the general context and compositional structure. A straightforward choice is to compute the anchor using the post-personalization text encoder $\tau_{\phi}$, i.e., $\tau_{\phi}(\lfloor p, c \rfloor)$. In practice, however, this anchor can be unreliable: learning the personalized token $V^*$ can also distort the embedding space around the base class concept, even when the class token $c$ is not explicitly optimized.

% \paragraph{Why $\tau_{\phi}$ can be a biased anchor.}
% Personalization aims to inject strong visual identity information into $V^*$. In practice, this update is not perfectly localized to the new token. Due to shared parameters and coupling in the text encoder (and its interaction with cross-attention), the learned concept can ``pull'' nearby semantic regions, altering the representations of the class token $c$ and semantically related concepts (e.g., ``kid'' when $c$ is ``man''). As a result, $\tau_{\phi}(\lfloor p, c \rfloor)$ may already carry traces of the personalized subject, effectively contaminating the anchor with identity-specific bias. When such a biased anchor is used, the residual term no longer isolates drift attributable to $V^*$; instead, the correction unintentionally mixes identity and context, which can degrade prompt fidelity and compositionality.
\vspace{-1em}
\begin{table}[h]

    \caption{Results on the CustomConcept101 dataset that conditioned on different text encoders. In which tort* means tortoise plushy, teddy* means plushy teddy bear, wpot* means wooden pot.
    % The \textcolor{blue}{blue number} indicates the proposed method outperforms its baseline counterpart, while the \textcolor{red}{red number} indicates the opposite. 
    % The GAP is the average improvement over all concepts. Qualitative results are shown in Fig. \ref{fig:qualitative_comparison_CC101}.
    }
    \label{tab:CC101_results_compare_tau}
    \centering
    \resizebox{0.9\textwidth}{!}{
    \begin{tabular}{lcccccccccc}
    \toprule
    Metric \quad \quad & dog & cat & tort* & teddy* & chair & table & flower & wpot* & barn & \quad Average\\
    \midrule

%     DB  & 20.28    &  20.29    &  24.55    &  18.89    &  21.87   &  20.77    &  20.07    &  20.43    &  23.58    &  21.19 \\
%    &  \rowcolor{lightgray!50}  59.93  &  74.82  &  91.14  &  85.52  &  87.50   &  73.83  &  82.98  &  66.69  &  80.98  & 78.15  \\
% DB+$TEA$  & 21.71    &  20.10    &  25.08    &  19.65    &  23.61    &  22.97    &  23.12    &  24.99    &  26.31    &  23.06  \\
%    &  \rowcolor{lightgray!50} 66.25  &  92.43  &  91.10  &  86.98   &  86.60  &  84.87  &  76.00  &  82.46  &  77.70 &   82.71 \\ \hdashline
    % \vspace{0.5em}
    \multicolumn{11}{l}{\multirow{2}{*}{$AdaptSP$ ($r_m$) with fine-tuned text encoder $\tau_\phi$ in Equation \ref{eq:second_decompose_rewrite}}} \\ 
    & \\
     CLIP-T$^p$ & 20.25 & 16.99 & 13.2 & 14.47 & 17.98 & 16.38 & 19.3 & 13.08 & 19.19 & 16.76 \\
    CLIP-T$^f$ & 26.09 & 22.58 & 25.45 & 22.74 & 23.63 & 24.31 & 23.23 & 23.37 & 26.76 & 24.24 \\
     \rowcolor{lightgray!50} CLIP-I &   66.28 & 88.92 & 88.42 & 89.97 & 80.98 & 80.14 & 74.83 & 86.6 & 80.45 & \textbf{81.84} \\
    \rowcolor{lightgray!50} DINO &48.69 & 79.29 & 73.86 & 49.93 & 68.00 & 73.37 & 50.36 & 80.97 & 61.57 & \textbf{65.12} \\ \midrule
      \multicolumn{11}{l}{\multirow{2}{*}{$AdaptSP$ ($r_m$) with original pre-trained text encoder $\tau_{\phi_0}$ in Equation \ref{eq:decompose_ori_text_en_rewrite}}} \\
      & \\
    CLIP-T$^p$ & 21.69 & 17.61 & 14.57 & 16.82 & 20.55 & 17.89 & 21.87 & 14.48 & 21.01 & \textbf{18.50} \\
    CLIP-T$^f$ & 27.34 & 23.4 & 26.81 & 25.18 & 26.22 & 25.49 & 25.37 & 24.56 & 27.48 & \textbf{25.76} \\
    \rowcolor{lightgray!50} CLIP-I  & 59.64 & 87.03 & 87.93 & 83.98 & 79.79 & 76.92 & 66.74 & 85.79 & 77.16 & 78.33  \\ 
    \rowcolor{lightgray!50} DINO  &  42.56 & 74.02 & 72.13 & 47.14 & 54.74 & 69.76 & 41.65 & 78.62 & 56.51 & 59.68 \\ 
    \bottomrule
    \end{tabular}
    }
    % \vspace{-1em}
\end{table}

To obtain a clean and stable anchor, we instead compute the class-prompt embedding using the original pre-trained text encoder $\tau_{\phi_0}$ ( Equation ~\ref{eq:decompose_ori_text_en_rewrite} and~\ref{eq:residual_projection_form_origin_tau}). Since $\tau_{\phi_0}$ is never exposed to the personalization objective, it provides a consistent representation of general language semantics and prompt composition. In AdaptSP, this stable anchor helps the residual term focus on the identity drift introduced by $V^*$, while preserving the contextual content in $p$, including objects, attributes, relations, and style descriptors, as intended.

% \paragraph{Effect of using $\tau_{\phi_0}$.}
We evaluate this design choice in Table~\ref{tab:CC101_results_compare_tau}. Specifically, we train DreamBooth on CustomConcept101 for 2000 steps and generate images using the same fine-tuned model, changing only the text encoder used to compute the class-prompt embedding: $\tau_{\phi}$ and $\tau_{\phi_0}$. The results show that using $\tau_{\phi_0}$ improves prompt fidelity (CLIP-T$^{f}$ score) while slightly reducing subject similarity measured by CLIP-I and DINO. This trade-off is expected: replacing $\tau_{\phi}$ with the stable anchor encourages the model to follow contextual tokens more faithfully and reduces training-specific biases (e.g., background/style correlations) that can inflate image-image similarity scores under overfitting.

\section{Conclusion and Future Work}
% \section{Conclusion}
% \label{sec:conclusion}

We studied the semantic collapsing problem in personalized text-to-image generation, where a learned concept token can dominate the prompt and suppress contextual details. Our analysis shows that this failure mode is associated with structured semantic drift: the difference between a personalized prompt and its prior class prompt forms residual embeddings that are (i) highly consistent across contexts and (ii) concentrated in a low-dimensional subspace.

Based on these observations, we proposed AdaptSP, a simple, training-free test-time mechanism that decouples identity (residual) from context (class-anchored prompt), augmenting prior global test-time adjustment approaches with subspace targeting.
AdaptSP can be plugged into existing personalization pipelines and improves prompt fidelity while maintaining strong subject identity, with a controllable trade-off between the two objectives. 

AdaptSP focuses specifically on adjusting prompt embeddings to mitigate SCP. This is also a limitation, as semantic collapse may arise from other components of the generation pipeline, such as the denoising network, cross-attention layers, or personalization training dynamics. Nevertheless, the idea of modeling structured residual drift is not limited to personalized text-to-image generation. It may also be useful for other embedding-based adaptation settings, including concept editing, controllable generation, representation steering, and multimodal alignment, where preserving a target concept while avoiding unintended semantic interference is essential. This opens a promising direction for future work on structured, training-free control of generative representations.

% We also discussed limitations of commonly used subject-fidelity metrics such as CLIP-I and DINO, particularly under overfitting, where high similarity scores may reflect background or style matching rather than robust identity preservation.

\clearpage

\bibliographystyle{plain}
\bibliography{main}

%%%%%%%%%%%%%%%%%%%%%%%%%%%%%%%%%%%%%%%%%%%%%%%%%%%%%%%%%%%%
\clearpage
\appendix

\section{Additional experiments and ablation studies}
\label{sec:appendix_experiments}

\subsection{Limitations of existing subject-fidelity metrics} 
\label{subsec:metric}

\begin{figure}[t]
    \centering
    \includegraphics[width=0.9\linewidth]{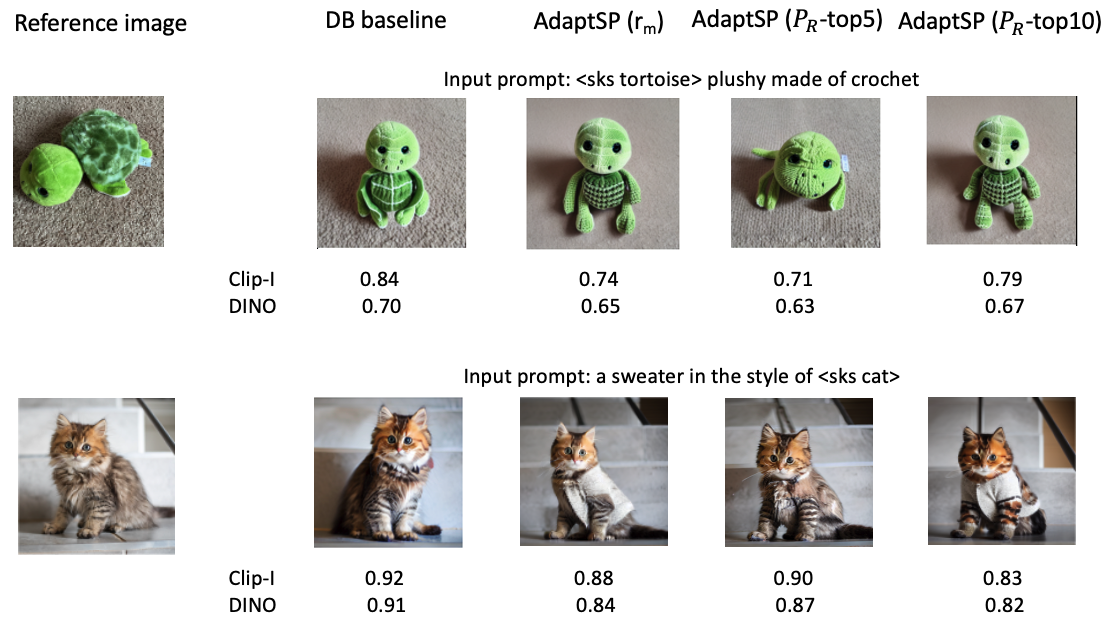}
    % \vspace{-1em}
    \caption{Problem of Subject Fidelity Metrics that can assign artificially high subject-fidelity scores to overfitting generations that reproduce training-specific details (e.g., background, color tone, texture patterns, and poses). In contrast, more diverse and context-generalized generations may receive lower scores despite preserving the subject identity.}
    \label{fig:metric_prob}
    \vspace{-1em}
\end{figure}

Existing subject-fidelity metrics in personalized generation commonly rely on pretrained visual encoders such as CLIP-I and DINO. CLIP-I is trained on large-scale text-image pairs with a contrastive objective that aligns image and text embeddings~\cite{radford2021learning,hessel2021clipscore}. As a result, its similarity tends to be driven by attributes that are easy to match to text descriptions (e.g., coarse appearance and high-level semantics), and it can be less sensitive to fine-grained identity cues that are rarely stated explicitly in captions~\cite{radford2021learning,hessel2021clipscore}. In contrast, DINO is trained in a self-supervised manner via knowledge distillation and encourages invariance across strong data augmentations, producing instance-level visual representations without text supervision~\cite{caron2021emerging}.

\begin{table}[h!]
\centering
\caption{Human evaluation result}
\label{tab:human_result}

\resizebox{0.9\textwidth}{!}{
\begin{tabular}{lccccccc}
\toprule
\multirow{2}{*}{Method}             & \multicolumn{3}{c}{\textbf{Subject   identity preservation}} &  & \multicolumn{3}{c}{\textbf{Prompt fidelity}}             \\  
                                     & Woman    & Cat          & Tortoise plushy   &  & Woman      & Cat        & Tortoise plushy \\ \midrule
\textbf{DreamBooth}                          &       \textbf{8.275}    & \textbf{9.980}    & \textbf{7.628}           &     & 8.251 & 5.000 & 6.103           \\
AdaptSP ($r_m$)                    & 7.888    & 9.298     & 6.761              &  &  \textbf{8.864} & \textbf{7.504} & \textbf{7.228}           \\
AdaptSP   ($R_{R}$-top5)             & 7.918    & 9.642     & 6.899              &   & 8.834 & 6.571 & 6.789           \\
AdaptSP   ($P_{\mathcal{R}}$-top10) &      7.978    & 9.683     & 6.954              &     & 8.714 & 6.479 & 6.883           \\
\midrule \midrule
\textbf{Custom   Diffusion}                     &      5.881    & \textbf{9.815}     & \textbf{6.983}              &    &    8.329   & 8.775 & 8.517           \\
AdaptSP ($r_m$)                  &    5.437      & 8.775     & 4.508              &   &   \textbf{8.575}    & \textbf{9.260} & \textbf{8.883}           \\
AdaptSP   ($R_{R}$-top5)            &   5.954       & 9.105     & 5.925              &  &   7.942   & 8.150 & 7.817           \\
AdaptSP   ($P_{\mathcal{R}}$-top10) &     \textbf{6.228}     & 9.065     & 6.192              &  &  7.927   & 8.040 & 8.267           \\
\bottomrule
\end{tabular}
}
\end{table}

Despite these strengths, both CLIP-I and DINO can be confounded by non-identity correlations in personalization datasets. Their similarity scores often increase when generated images match the reference set in low-level or contextual factors, such as background, color palette, lighting, texture patterns, viewpoint, and typical subject poses \cite{ruiz2023dreambooth}. This leads to a critical failure mode: when a method overfits (e.g., memorizes training images or their scene statistics), these metrics may report very high subject-fidelity scores because the model reproduces similar environments and visual style, even if it fails to generalize the subject identity to novel contexts. Therefore, high CLIP-I/DINO similarity does not necessarily indicate robust identity preservation; it may instead reflect background and style matching. We present this problem with some samples in Figure \ref{fig:metric_prob}.

To better assess quality beyond automatic metrics, we conduct an additional human evaluation. Each participant is asked to rate the generated images on a scale from 0 to 10 for both subject identity preservation and prompt fidelity. Higher scores indicate better preservation of the target subject and stronger consistency with the given prompt. The results are provided in Table \ref{tab:human_result}. Overall,

\subsection{Result on CC101 dataset}

% \vspace{-2em}

\begin{table}[h!]
    \caption{Results on CustomConcept101 dataset, tort* means tortoise plushy, teddy* means plushy teddy bear, wpot* means wooden pot. The first/\colorbox{lightgray!50} {second} metric is the $\text{CLIP-T}^{f}$/\colorbox{lightgray!50}{CLIP-I} score. The highest score is shown in \textbf{bold}, and the runner-up is \underline{underlined}.
    % The \textcolor{blue}{blue number} indicates the proposed method outperforms its baseline counterpart, while the \textcolor{red}{red number} indicates the opposite. 
    % The GAP is the average improvement over all concepts. Qualitative results are shown in Fig. \ref{fig:qualitative_comparison_celeba}.
    }
    \label{tab:cc101_results}
    \centering
    \resizebox{1\textwidth}{!}{
    \begin{tabular}{lcccccccccc}
    \toprule
    Method &  dog & cat & tort* & teddy* & chair & table & flower & wpot* & barn & Average \\
    \midrule
    % Textual Inversion  & 23.70    &  26.29    &  21.32    &  26.06    &  25.25    &  23.46    &  26.54    &  24.61   &  25.40    &  24.97  \\
    %     & \rowcolor{lightgray!50}  68.93  &  61.68  &  65.33  &  75.64  &  69.63  &  71.19  &  74.37  &  64.50 &  67.55  &  60.88  \\
    % Textual Inversion+$TEA$  & 24.23    &  26.49   & 20.97   &  27.28    &  26.11    &  23.93    &  26.89    &  25.98    &  25.69    &  25.68 \\
    %          & \rowcolor{lightgray!50}  68.48  &  59.60  &  65.20   &  71.01  &  65.73  &  66.25  &  70.56  &  62.97  &  66.72  &  59.01  \\
    % \midrule
    DreamBooth  & 20.97 & 19.21 & 25.43 & 20.9 & 21.44 & 21.57 & 20.19 & 23.14 & 24.54 & 21.93  \\
 \rowcolor{lightgray!50} \cellcolor{white} &   70.19 & 93.49 & 90.06 & 92.17 & 84.8 & 85.65 & 82.29 & 87.55 & 82.81 & \textbf{85.45} \\
    DreamBooth+$TEA$  & 21.71    &  20.10    &  25.08    &  19.65    &  23.61    &  22.97    &  23.12    &  24.99    &  26.31    &  23.06  \\
 \rowcolor{lightgray!50} \cellcolor{white} & 66.25  &  92.43  &  91.10  &  86.98   &  86.60  &  84.87  &  76.00  &  82.46  &  77.70 &   \underline{82.71} \\ \cdashlinelr{1-11}

    DreamBooth+$AdaptSP$ \\
    \quad \quad ($r_m$)  &   27.34 & 23.4 & 26.81 & 25.18 & 26.22 & 25.49 & 25.37 & 24.56 & 27.48 & \textbf{25.76}     \\
 \rowcolor{lightgray!50} \cellcolor{white} &59.64 & 87.03 & 87.93 & 83.98 & 79.79 & 76.92 & 66.74 & 85.79 & 77.16 & 78.33 \\
    \quad \quad ($P_{\mathcal{R}}$)  &   26.16 & 21.74 & 26.22 & 24.41 & 24.8 & 23.02 & 23.98 & 23.69 & 26.21 & \underline{24.47}     \\
 \rowcolor{lightgray!50} \cellcolor{white} & 60.37 & 87.16 & 89.21 & 83.66 & 81.2 & 80.78 & 68.72 & 86.39 & 79.15 & 79.63 \\ \midrule
     
    Class Diffusion  & 26.98 & 27.41 & 29.71 & 26.08 & 25.88 & 27.1 & 25.49 & 28.12 & 27.7 & 27.16    \\
 \rowcolor{lightgray!50} \cellcolor{white} & 61.17 & 82.13 & 78.97 & 86.00 & 70.53 & 64.39 & 67.31 & 73.32 & 76.22 & \textbf{73.34} \\

Class Diffusion+$TEA$  & 27.72 & 28.12 & 29.00 & 26.07 & 26.94 & 27.16 & 27.21 & 28.51 & 27.65 & \underline{27.59}     \\
 \rowcolor{lightgray!50} \cellcolor{white} & 56.44 & 77.26 & 78.87 & 77.46 & 75.56 & 60.54 & 64.87 & 60.72 & 71.27 & 69.22  \\ 
\cdashlinelr{1-11}
    Class Diffusion+$AdaptSP$ \\
    \quad \quad ($r_m$) & 27.30 & 29.06 & 29.79 & 26.18 & 25.88 & 27.10 & 26.60 & 28.82 & 28.52 & \textbf{27.69}\\
 \rowcolor{lightgray!50} \cellcolor{white} & 57.10 & 77.05 & 72.25 & 85.70 & 67.92 & 59.56 & 63.67 & 67.07 & 70.02 & 68.93  \\

    \quad \quad ($P_{\mathcal{R}}$) & 26.88 & 26.05 & 28.98 & 25.55 & 25.08 & 25.61 & 25.43 & 28.29 & 26.55 & 26.49 \\
 \rowcolor{lightgray!50} \cellcolor{white} & 60.22 & 80.89 & 78.24 & 85.82 & 69.95 & 64.02 & 64.82 & 70.61 & 74.55 & \underline{72.12}  \\ \midrule
       
    Custom Diffusion  & 26.81 & 27.65 & 29.47 & 26.27 & 25.5 & 26.32 & 25.02 & 28.0 & 26.9 & 26.88   \\
 \rowcolor{lightgray!50} \cellcolor{white} & 61.39 & 81.24 & 78.76 & 84.61 & 69.62 & 61.36 & 67.29 & 68.83 & 76.73 & 72.20 \\

Custom Diffusion+$TEA$  & 27.45    &  29.25    &  31.06    &  28.04    &  28.16    &  28.34    &  27.37    &  30.15   &  27.82    & \textbf{28.63}   \\
        &  56.18  &  77.59  &  79.61  &  78.64  &  71.81  &  63.08  &  62.63  &  68.0  &  71.78  & 69.92 \\ \cdashlinelr{1-11}
    Custom Diffusion+$AdaptSP$ \\
    \quad \quad ($r_m$) & 27.42 & 29.2 & 30.22 & 26.6 & 25.53 & 26.65 & 26.44 & 28.96 & 28.6 & \underline{27.74} \\
 \rowcolor{lightgray!50} \cellcolor{white} & 57.1 & 77.04 & 73.47 & 85.09 & 67.1 & 58.06 & 63.49 & 67.76 & 69.65 & 68.75  \\

    \quad \quad ($P_{\mathcal{R}}$) & 26.58 & 25.67 & 29.37 & 25.85 & 25.37 & 25.93 & 25.4 & 28.0 & 26.32 & 26.50 \\
 \rowcolor{lightgray!50} \cellcolor{white} & 59.29 & 81.3 & 78.79 & 85.59 & 68.87 & 63.3 & 64.89 & 73.44 & 77.32 & \textbf{72.53}  \\
    \bottomrule
    \end{tabular}
    }
    % \vspace{-2em}
\end{table}

As mentioned in Section 4, we present a detailed result on the CC101 dataset in Table \ref{tab:cc101_results}.

\subsection{Experiments on Flux pre-trained model}

Pre-trained FLUX models provide strong image generation quality and prompt fidelity. Building on these strengths, several personalization methods have been developed on top of FLUX, such as EasyControl \cite{zhang2025easycontrol} and OminiControl \cite{tan2025ominicontrol, tan2025ominicontrol2}. These methods introduce additional lightweight LoRA modules to adapt the model for subject control in a parameter-efficient manner, aiming to generalize to new subjects without requiring full fine-tuning.

Although such methods can follow reference visual concepts to generate new images, they often fail to preserve the distinctive details that differentiate subjects within the same class. As a result, additional subject-specific fine-tuning is still needed to achieve faithful personalization.

In this section, we fine-tune FLUX following the EasyControl framework to obtain personalized models. We then apply TEA and AdaptSP ($r_m$ and $P_{\mathcal{R}}$-top2) only to the pooled prompt embedding and compare them with the EasyControl baseline. We observe that the pooled prompt embedding has a more limited influence on generation than the token-level prompt embeddings in FLUX. As a result, both TEA and AdaptSP yield modest improvements in prompt fidelity while largely preserving subject identity compared to the EasyControl baseline.

% \begin{table}[h!]
% \caption{Comparison of average performance on CelebA and CC101 when integrating AdaptSP with EasyControl on the Flux pre-trained models.
%     }
%     \label{tab:all_result}
%     \centering
%     \resizebox{1\textwidth}{!}{
% \begin{tabular}{cccccccccc}
% \toprule
%            & \multicolumn{4}{c}{\multirow{2}{*}{\textbf{CelebA dataset}} }                             &  & \multicolumn{4}{c}{\multirow{2}{*}{\textbf{CC101 dataset}}}                              \\ 
%            & \\
%            & \quad EasyControl        & \quad TEA & \quad AdaptSP ($r_m$)  & \quad AdaptSP ($P_{\mathcal{R}}$) & \quad \quad  & EasyControl       & \quad TEA & \quad AdaptSP ($r_m$) & \quad AdaptSP ($P_{\mathcal{R}}$) \\ \midrule \midrule
% \multicolumn{10}{l}{\multirow{2}{*}{\textbf{DreamBooth}}}               \\                                                                                                    \\
% Clip-T$^p$ & 19.41         & \textbf{19.46}   & - & 19.44                      &  & -   & -     & -   & -                      \\
% Clip-T$^f$ & 24.81      &  24.89    & - & \textbf{24.94  }    &  & -   & -     & -   & -               \\ \cdashlinelr{1-10}
% \rowcolor{lightgray!50} Clip-I     & 69.94     &    70.06   & -  & 69.93          &  & -   & -     & -   & -                  \\
% \rowcolor{lightgray!50} DINO       & \textbf{0.651} &  0.650 & -          & 0.650                      &  & -   & -     & -   & -                        \\ \midrule
% \bottomrule
% \end{tabular}
% }
% \end{table}

\begin{table}[h!]
\caption{Comparison of average performance on CelebA and CC101 when integrating AdaptSP with EasyControl on the Flux pre-trained models.
    }
    \label{tab:all_result}
    \centering
    % \resizebox{1\textwidth}{!}{
\begin{tabular}{ccccc}
\toprule
           & \multicolumn{4}{c}{\multirow{2}{*}{\textbf{CelebA dataset}} }  \\ 
           & \\
           & \quad EasyControl        & \quad TEA  & \quad AdaptSP ($P_{\mathcal{R}}$) \\ \midrule \midrule
% \multicolumn{5}{l}{\multirow{2}{*}{\textbf{DreamBooth}}}               \\                                                                                                    \\
Clip-T$^p$ & 19.41         & \textbf{19.46}  & 19.44                  \\
Clip-T$^f$ & 24.81      &  24.89   & \textbf{24.94  }  \\ \cdashlinelr{1-5}
\rowcolor{lightgray!50} Clip-I     & 69.94     &    70.06     & 69.93           \\
\rowcolor{lightgray!50} DINO       & \textbf{0.651} &  0.650      & 0.650 \\ \midrule
\bottomrule
\end{tabular}
% }
\end{table}

\subsection{Visualization of generated images}

\begin{figure}
    \centering
    \includegraphics[width=0.95\linewidth]{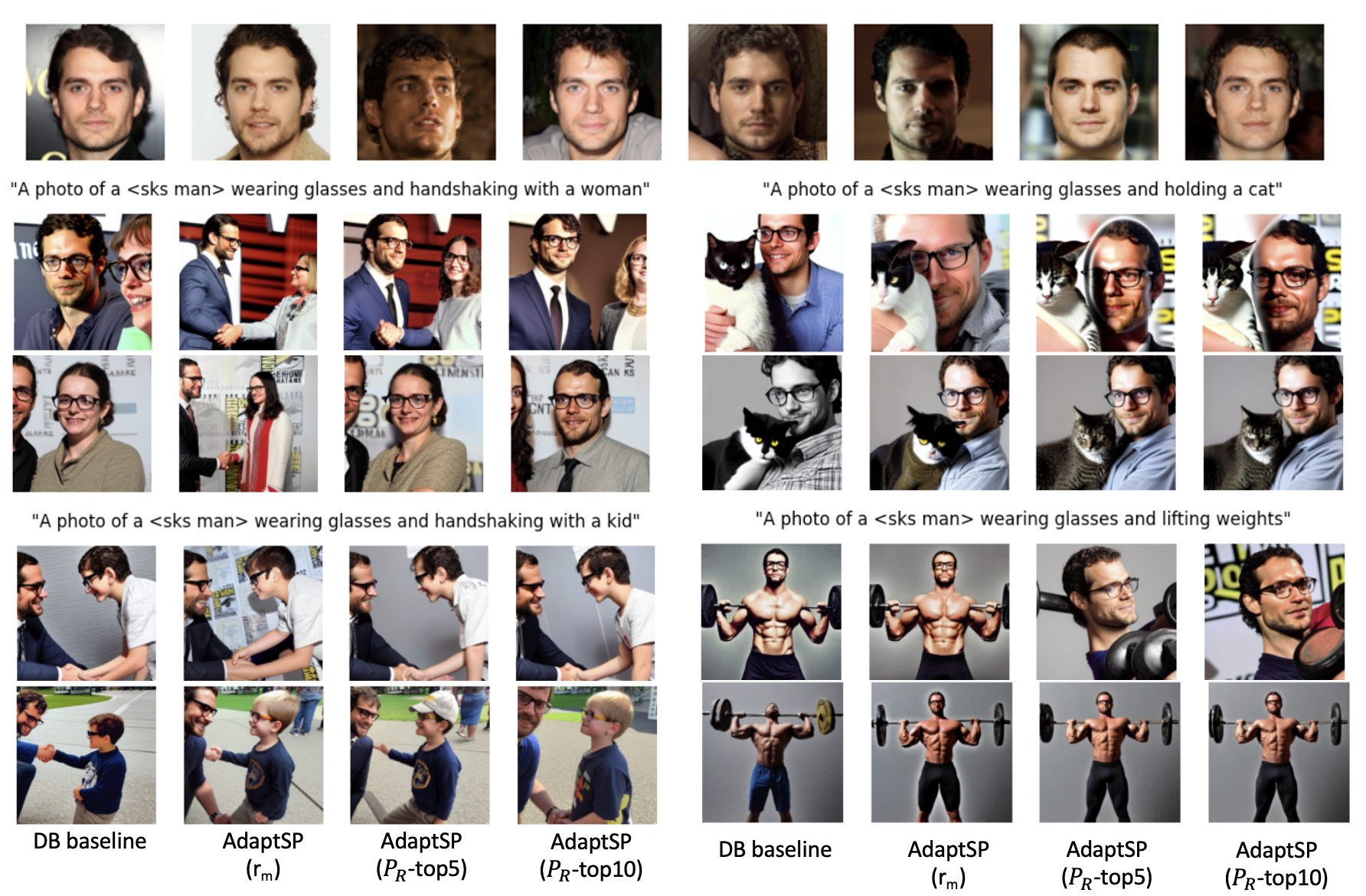} 
    \caption{\textbf{DreamBooth on CelebA (concept 342):} Qualitative comparison of DreamBooth and AdaptSP variants. The first row shows the reference images. The remaining rows show generations from different prompts and random seeds for each method.}
    \label{fig:db_celebA_342}
\end{figure}

\begin{figure}
    \centering
    \includegraphics[width=0.95\linewidth]{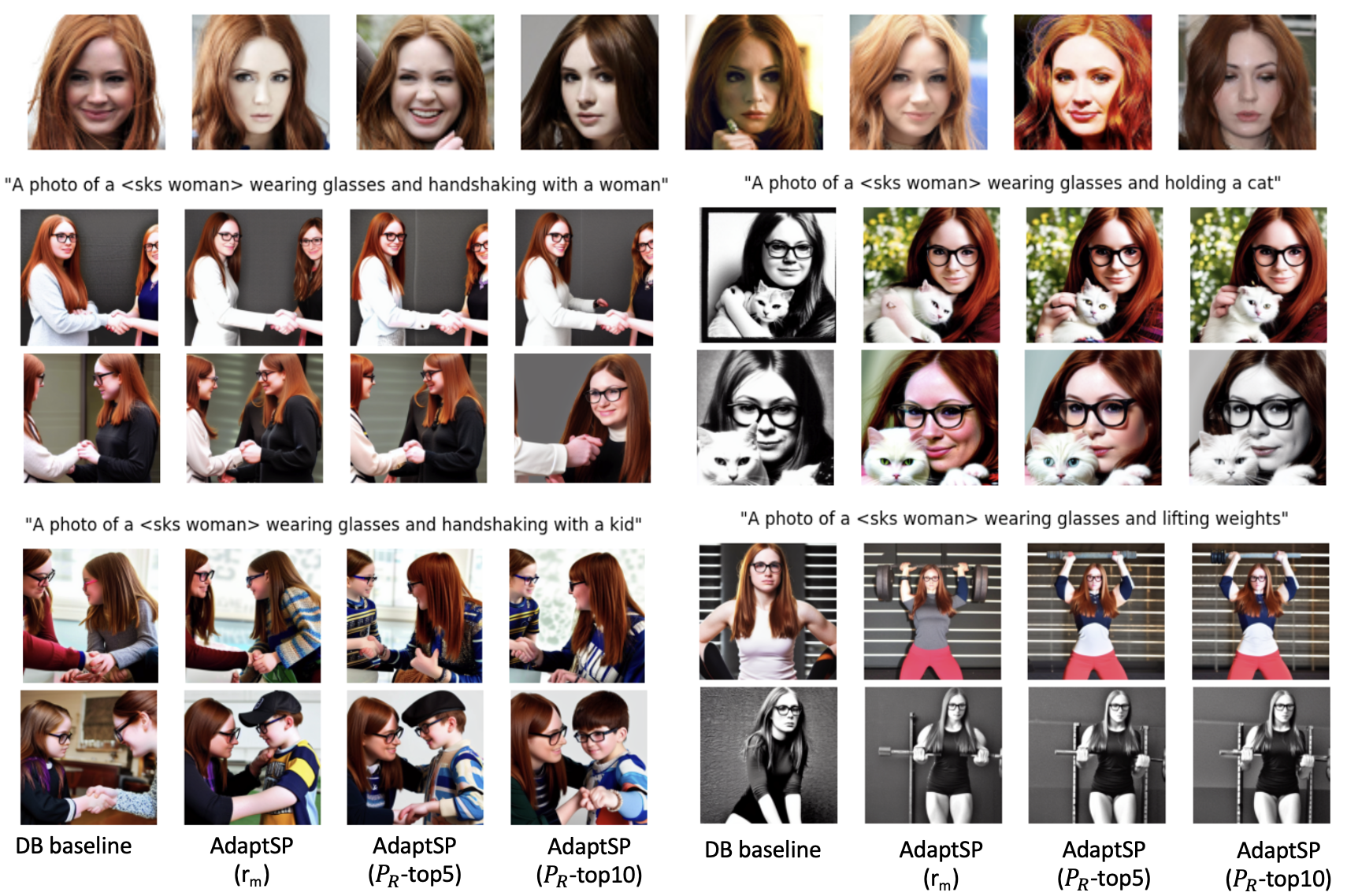}
    \caption{\textbf{DreamBooth on CelebA (concept 908):} Qualitative comparison of DreamBooth and AdaptSP variants. The first row shows the reference images. The remaining rows show generations from different prompts and random seeds for each method.}
    \label{fig:db_celebA_908}
\end{figure}

\begin{figure}
    \centering
    \includegraphics[width=0.95\linewidth]{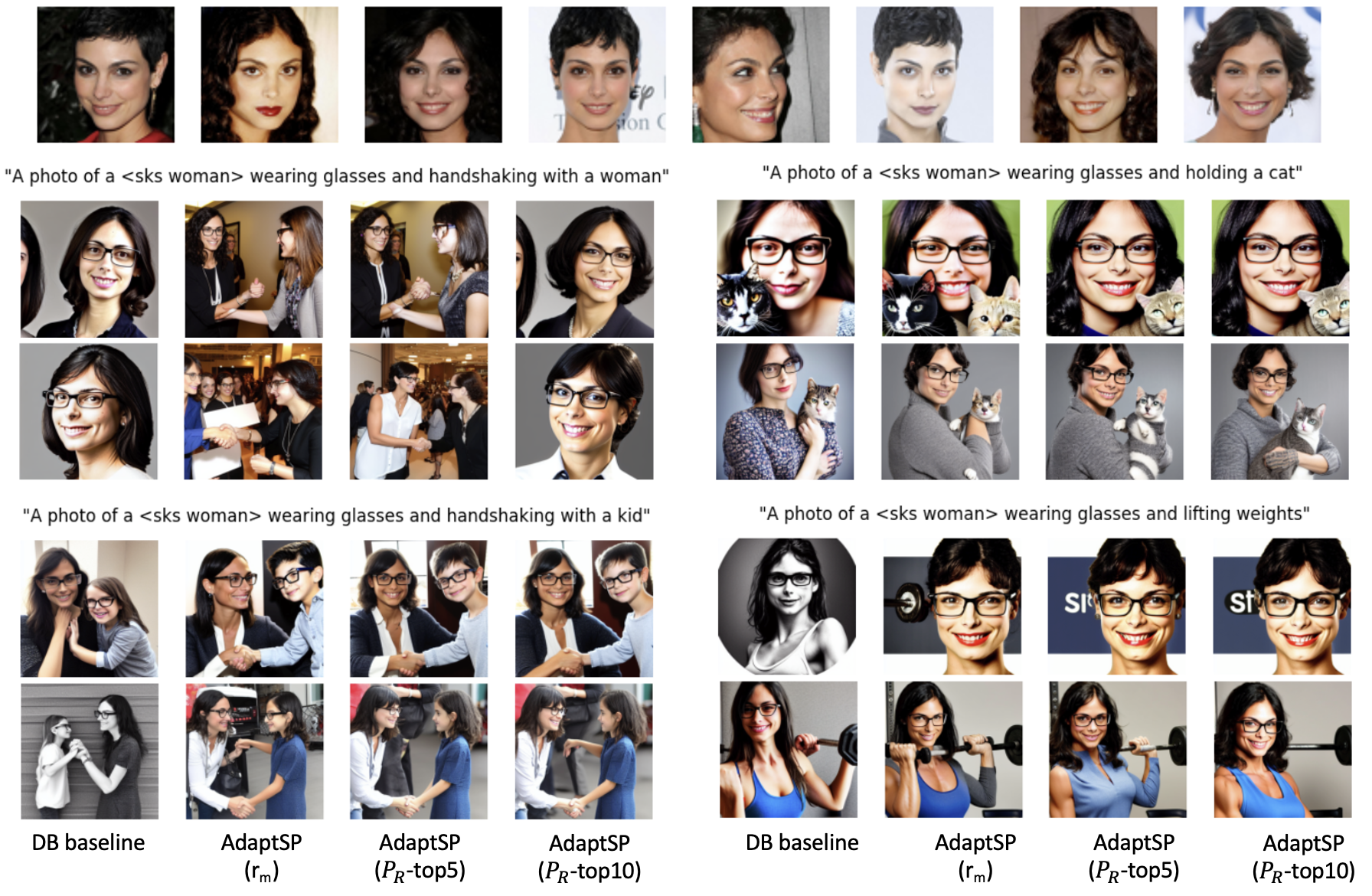}
    \caption{\textbf{DreamBooth on CelebA (concept 181):} Qualitative comparison of DreamBooth and AdaptSP variants. The first row shows the reference images. The remaining rows show generations from different prompts and random seeds.}
    \label{fig:db_celebA_181}
\end{figure}

\begin{figure}
    \centering
    \includegraphics[width=0.95\linewidth]{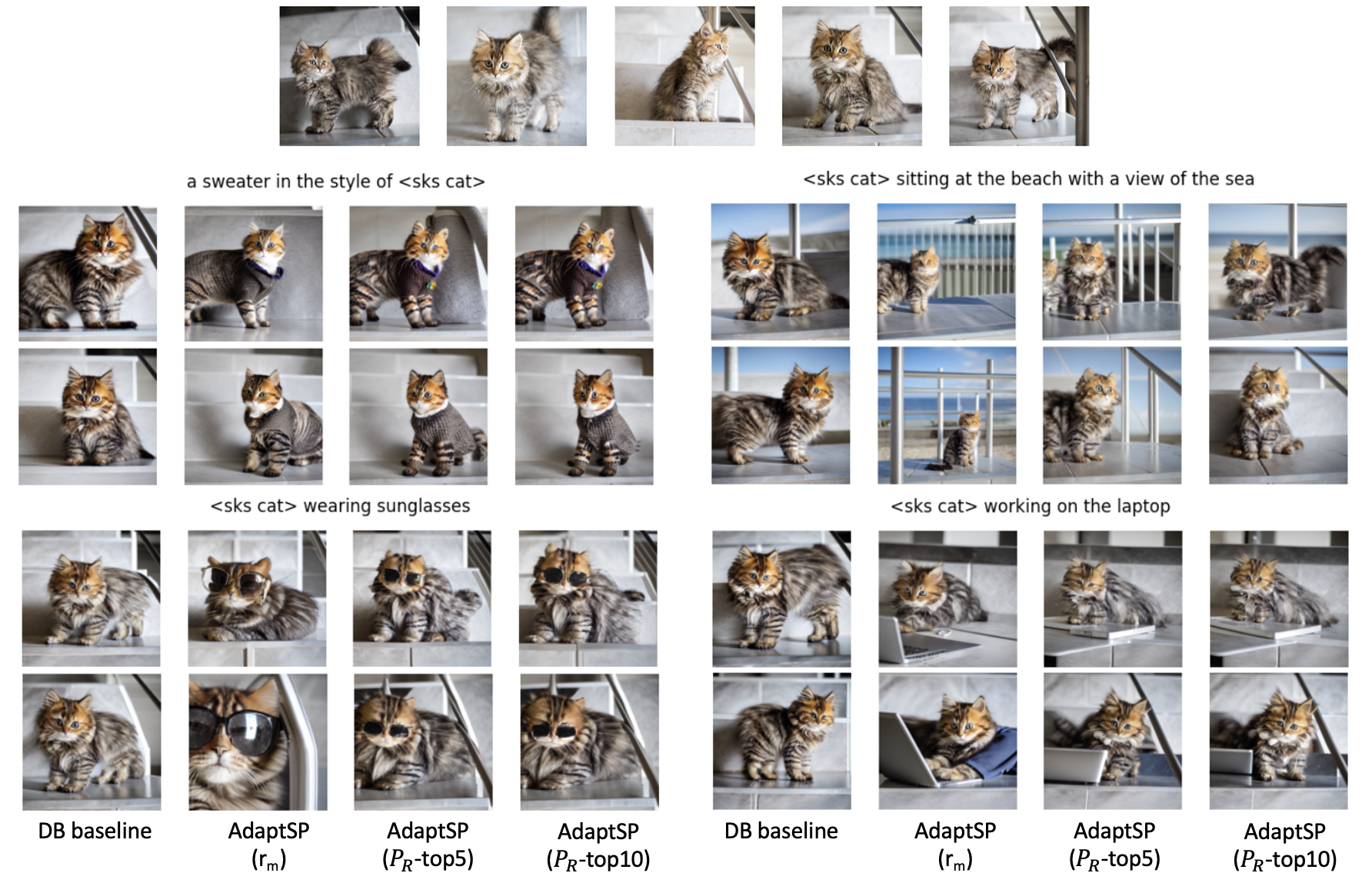}
    \caption{\textbf{DreamBooth on CC101 (concept cat):} Qualitative comparison of DreamBooth and AdaptSP variants. The first row shows the reference images. The remaining rows show generations from different prompts and random seeds.}
    \label{fig:db_cc101_cat}
\end{figure}

\begin{figure}
    \centering
    \includegraphics[width=0.95\linewidth]{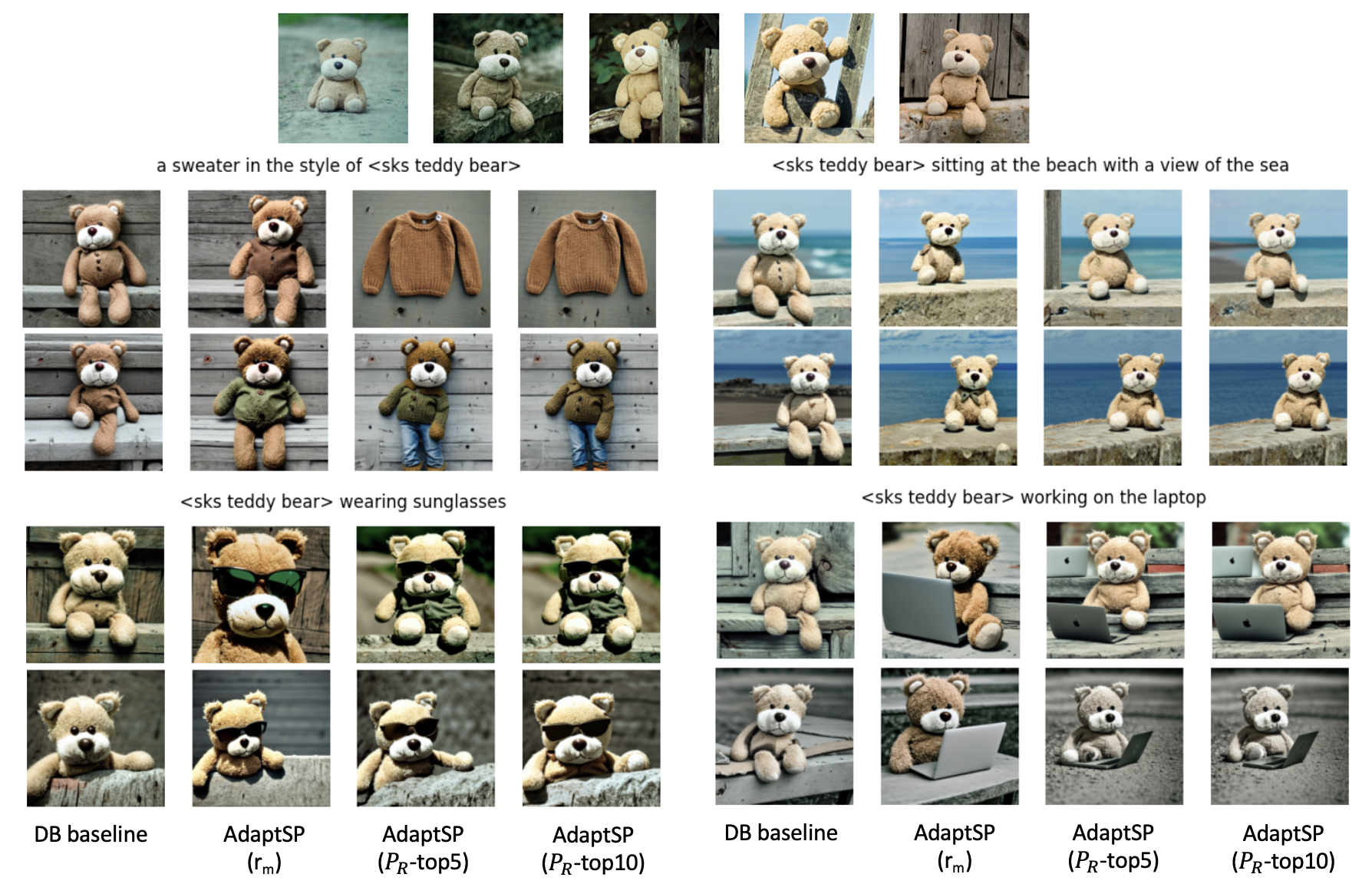}
    \caption{\textbf{DreamBooth on CC101 (concept teddy bear):} Qualitative comparison of DreamBooth and AdaptSP variants. The first row shows the reference images. The remaining rows show generations from different prompts and random seeds.}
    \label{fig:db_cc101_teddy}
\end{figure}

\begin{figure}
    \centering
    \includegraphics[width=0.95\linewidth]{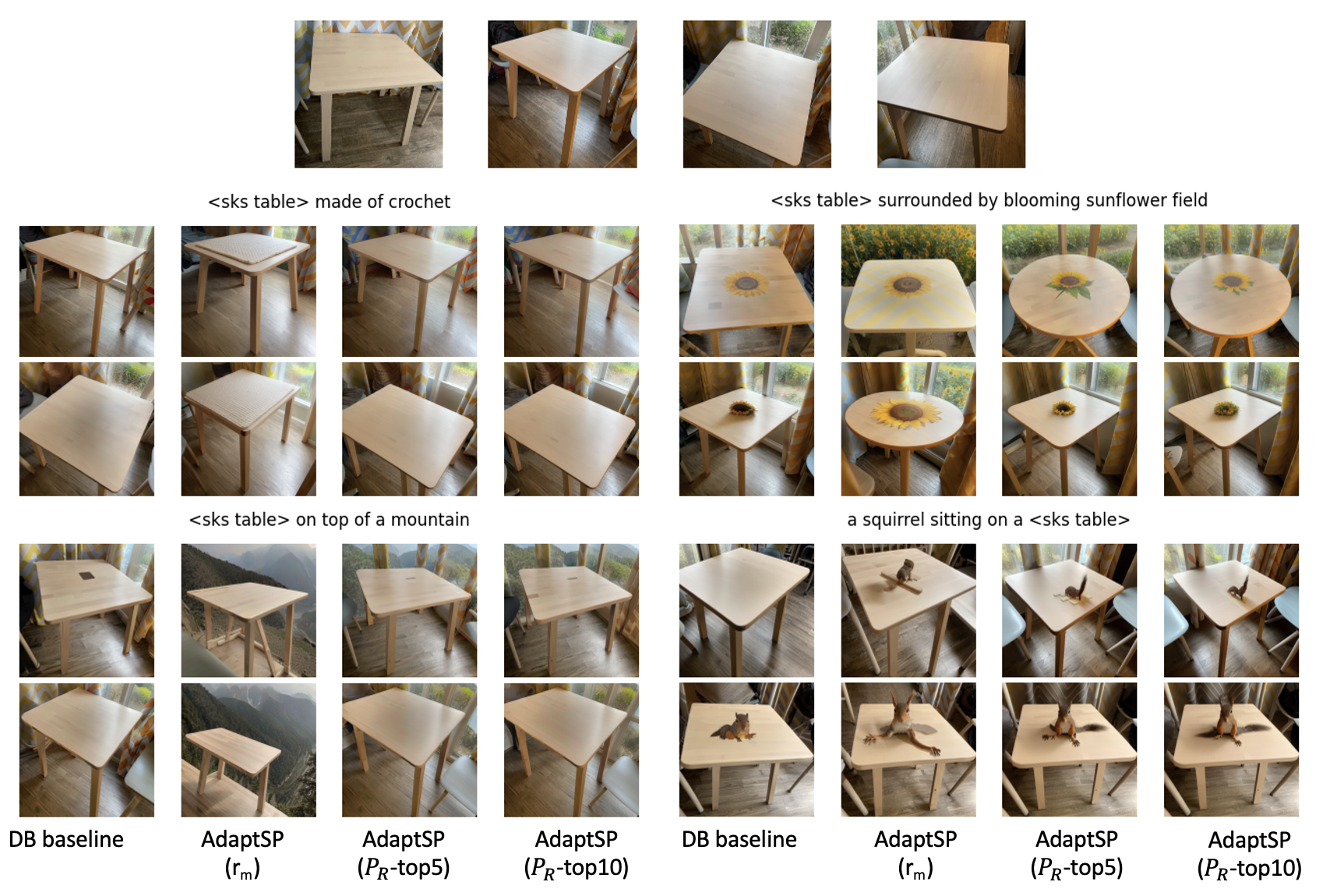}
    \caption{\textbf{DreamBooth on CC101 (concept table):} Qualitative comparison of DreamBooth and AdaptSP variants. The first row shows the reference images. The remaining rows show generations from different prompts and random seeds.}
    \label{fig:db_cc101_table}
\end{figure}

\begin{figure}
    \centering
    \includegraphics[width=0.95\linewidth]{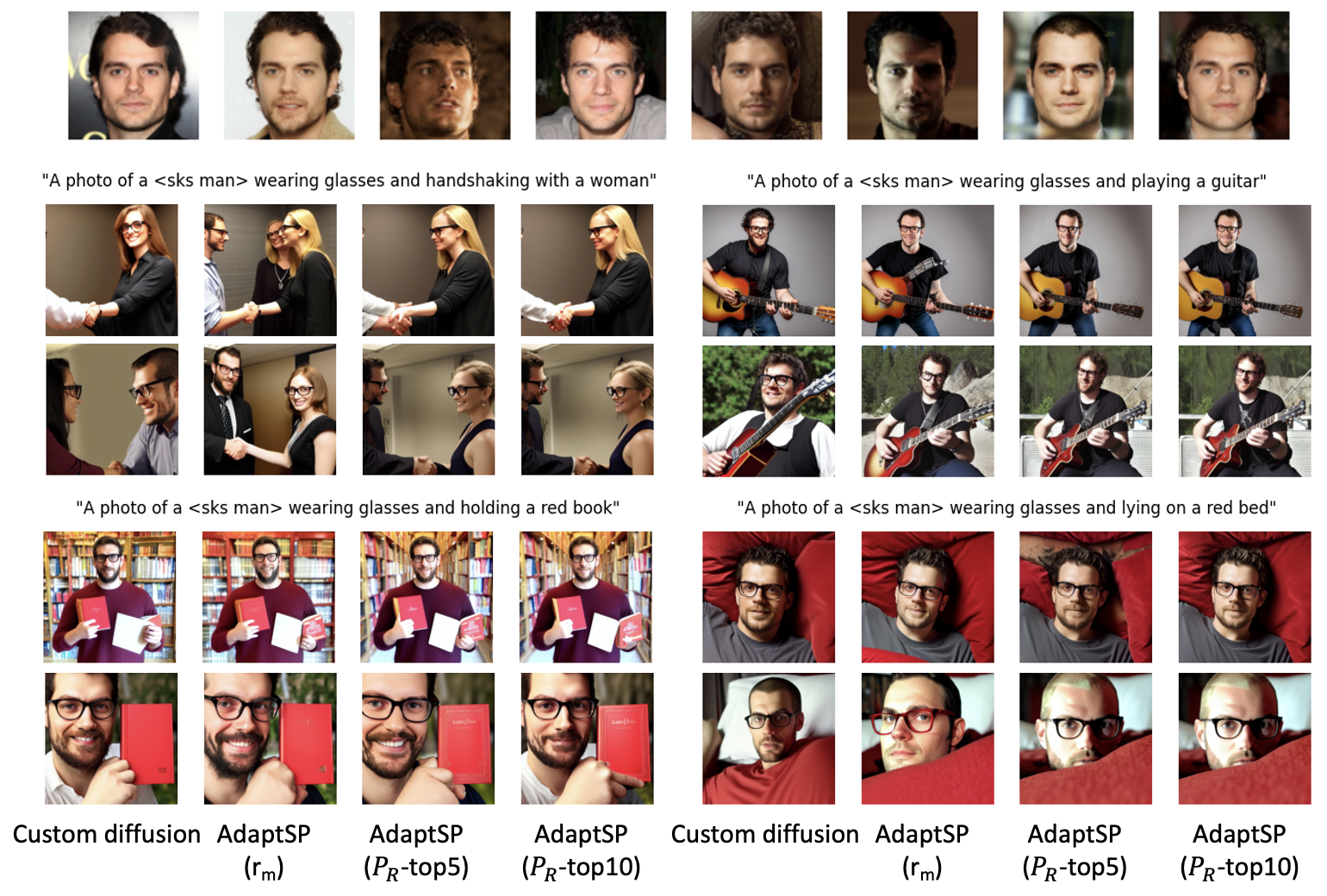}
    \caption{\textbf{Custom Diffusion on CelebA (concept 342):} Qualitative comparison of Custom Diffusion and AdaptSP variants. The first row shows the reference images. The remaining rows show generations from different prompts and random seeds.}
    \label{fig:cs_celebA_342}
\end{figure}

\begin{figure}
    \centering
    \includegraphics[width=0.95\linewidth]{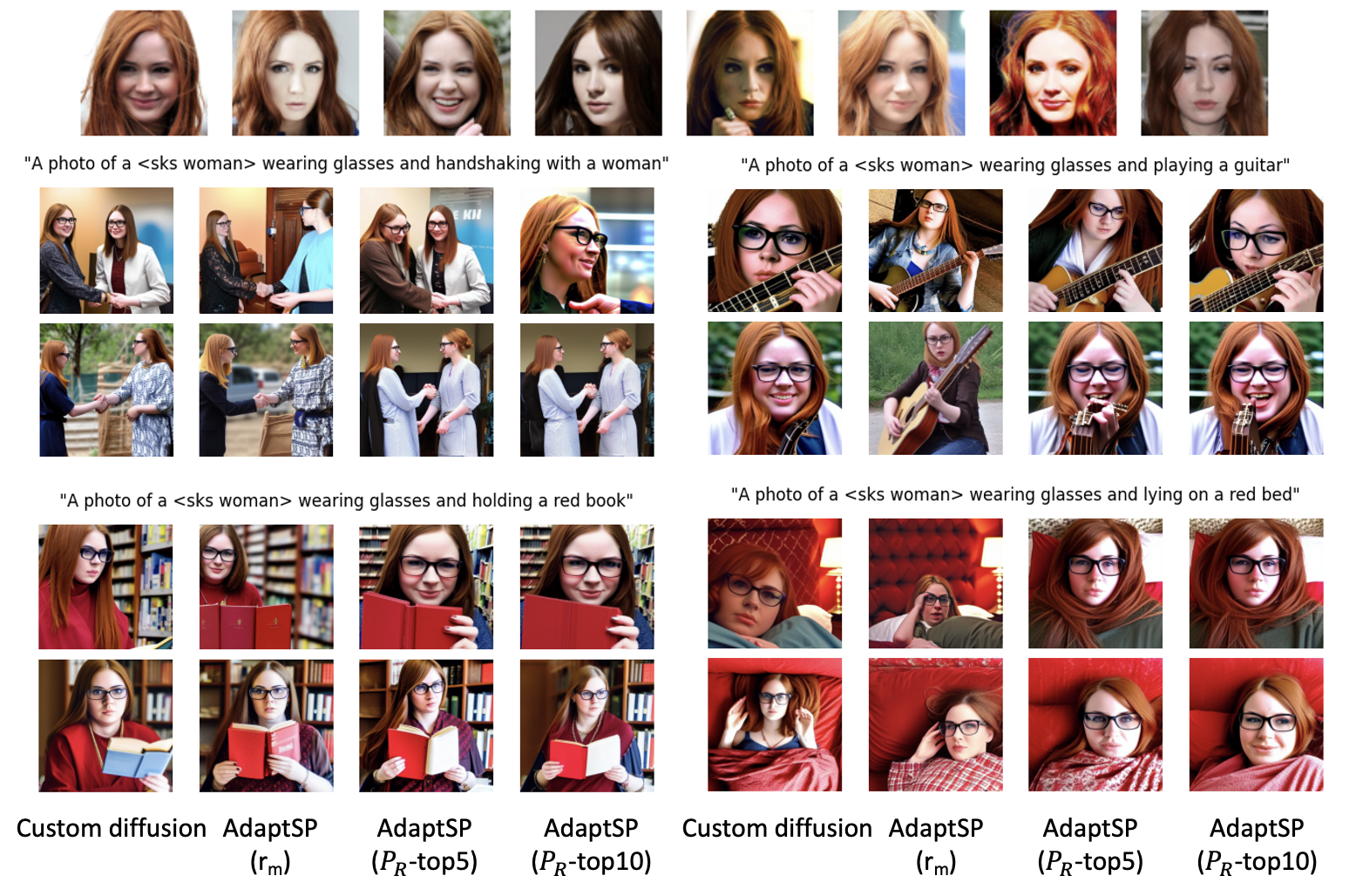}
    \caption{\textbf{Custom Diffusion on CelebA (concept 908):} Qualitative comparison of Custom Diffusion and AdaptSP variants. The first row shows the reference images. The remaining rows show generations from different prompts and random seeds.}
    \label{fig:cs_celebA_908}
\end{figure}

\begin{figure}
    \centering
    \includegraphics[width=0.95\linewidth]{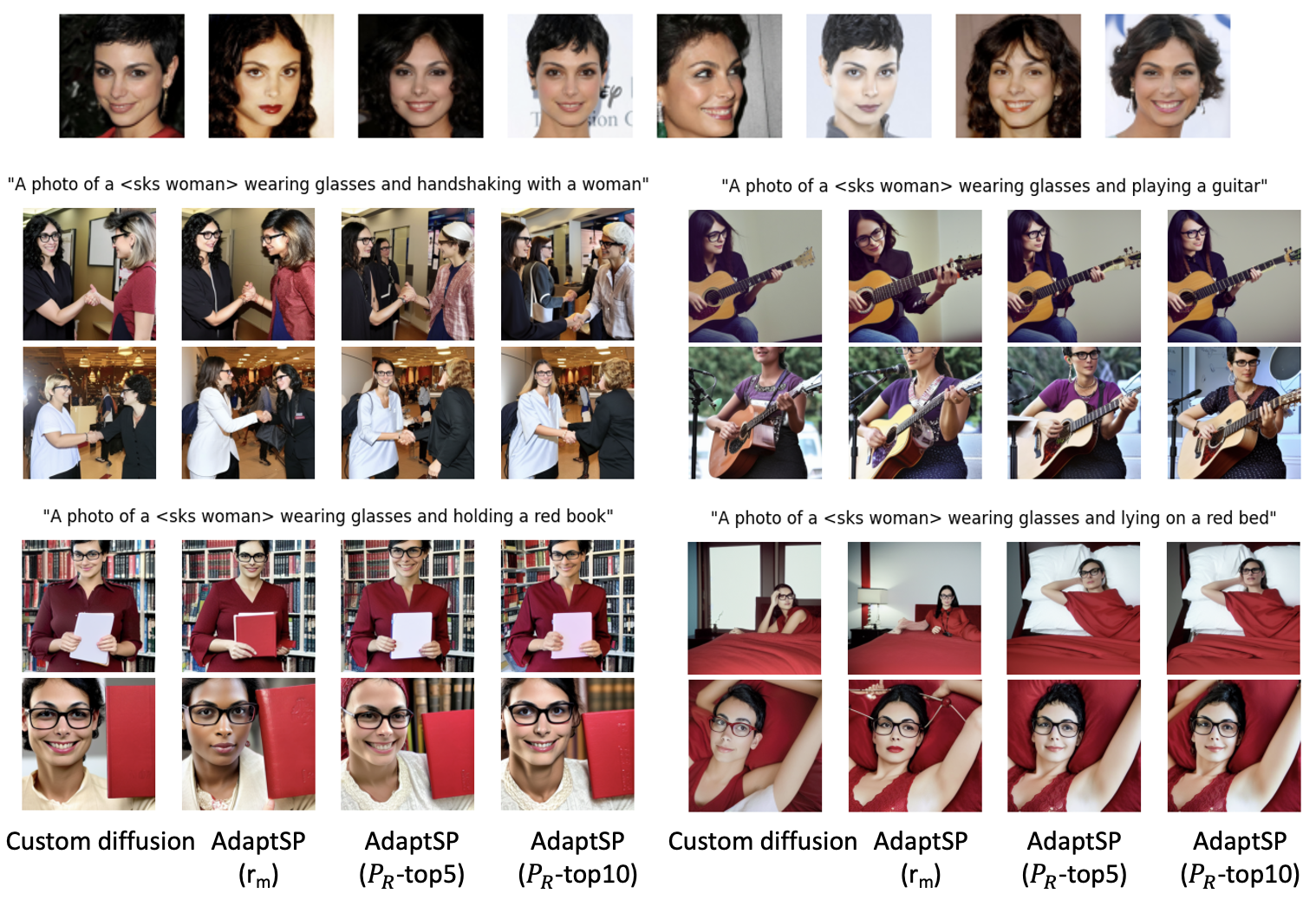}
    \caption{\textbf{Custom Diffusion on CelebA (concept 181):} Qualitative comparison of Custom Diffusion and AdaptSP variants. The first row shows the reference images. The remaining rows show generations from different prompts and random seeds.}
    \label{fig:cs_celebA_181}
\end{figure}

\begin{figure}
    \centering
    \includegraphics[width=0.95\linewidth]{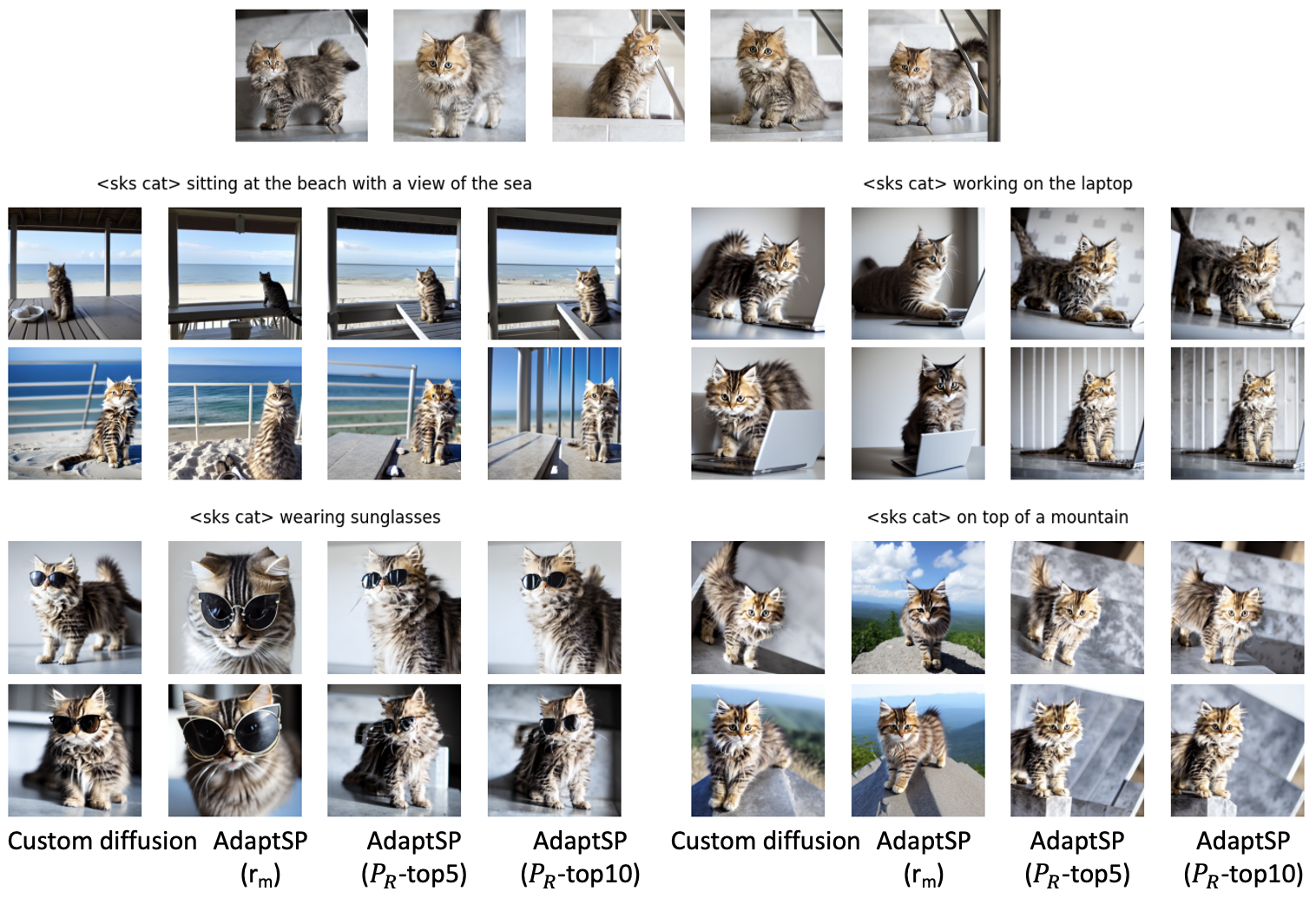}
    \caption{\textbf{Custom Diffusion on CC101 (concept cat):} Qualitative comparison of Custom Diffusion and AdaptSP variants. The first row shows the reference images. The remaining rows show generations from different prompts and random seeds.}
    \label{fig:cs_cc101_cat}
\end{figure}

\begin{figure}
    \centering
    \includegraphics[width=0.95\linewidth]{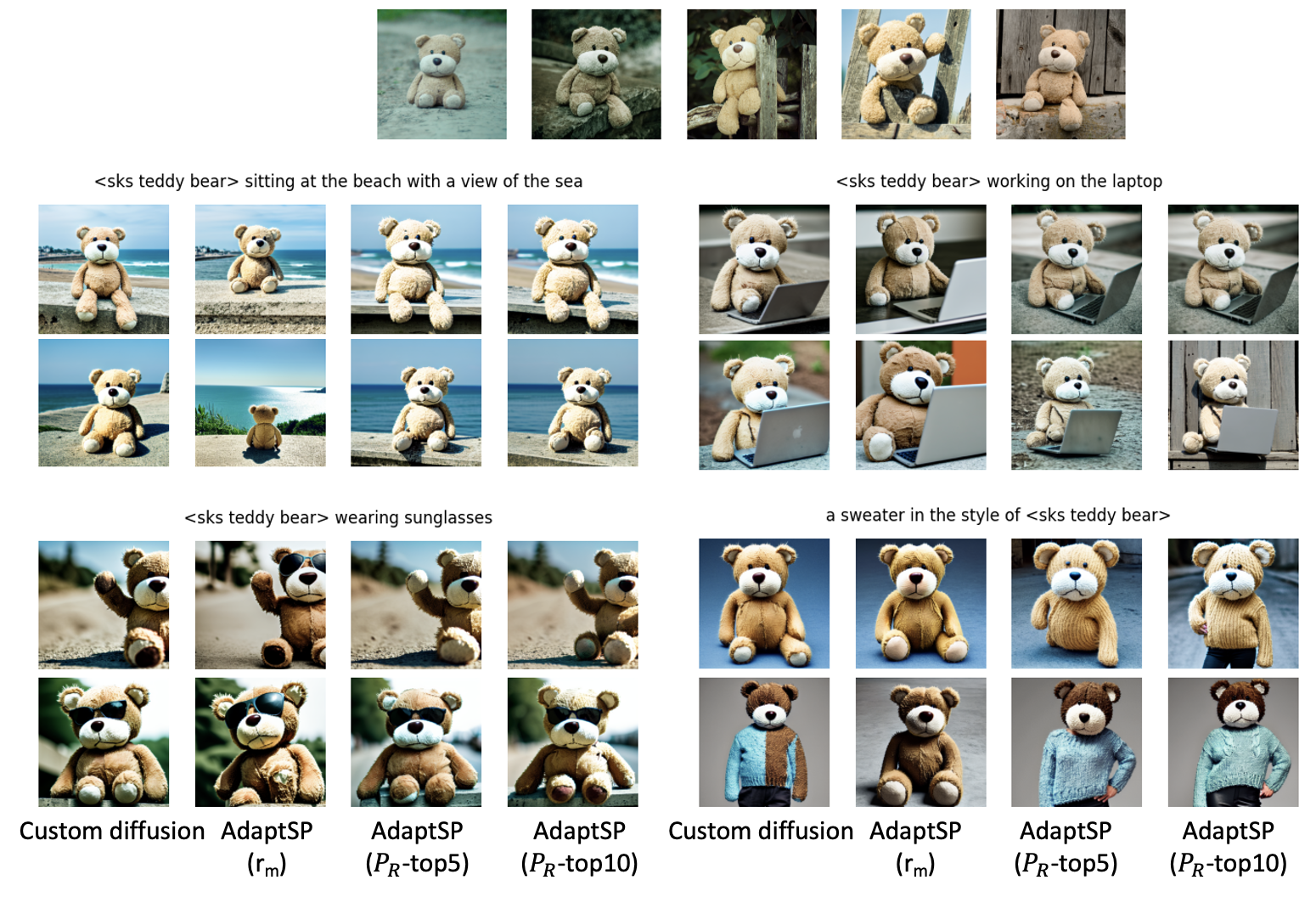}
    \caption{\textbf{Custom Diffusion on CC101 (concept teddy bear):} Qualitative comparison of Custom Diffusion and AdaptSP variants. The first row shows the training images. The remaining rows show generations from different prompts and random seeds.}
    \label{fig:cs_cc101_teddy}
\end{figure}

\begin{figure}
    \centering
    \includegraphics[width=0.95\linewidth]{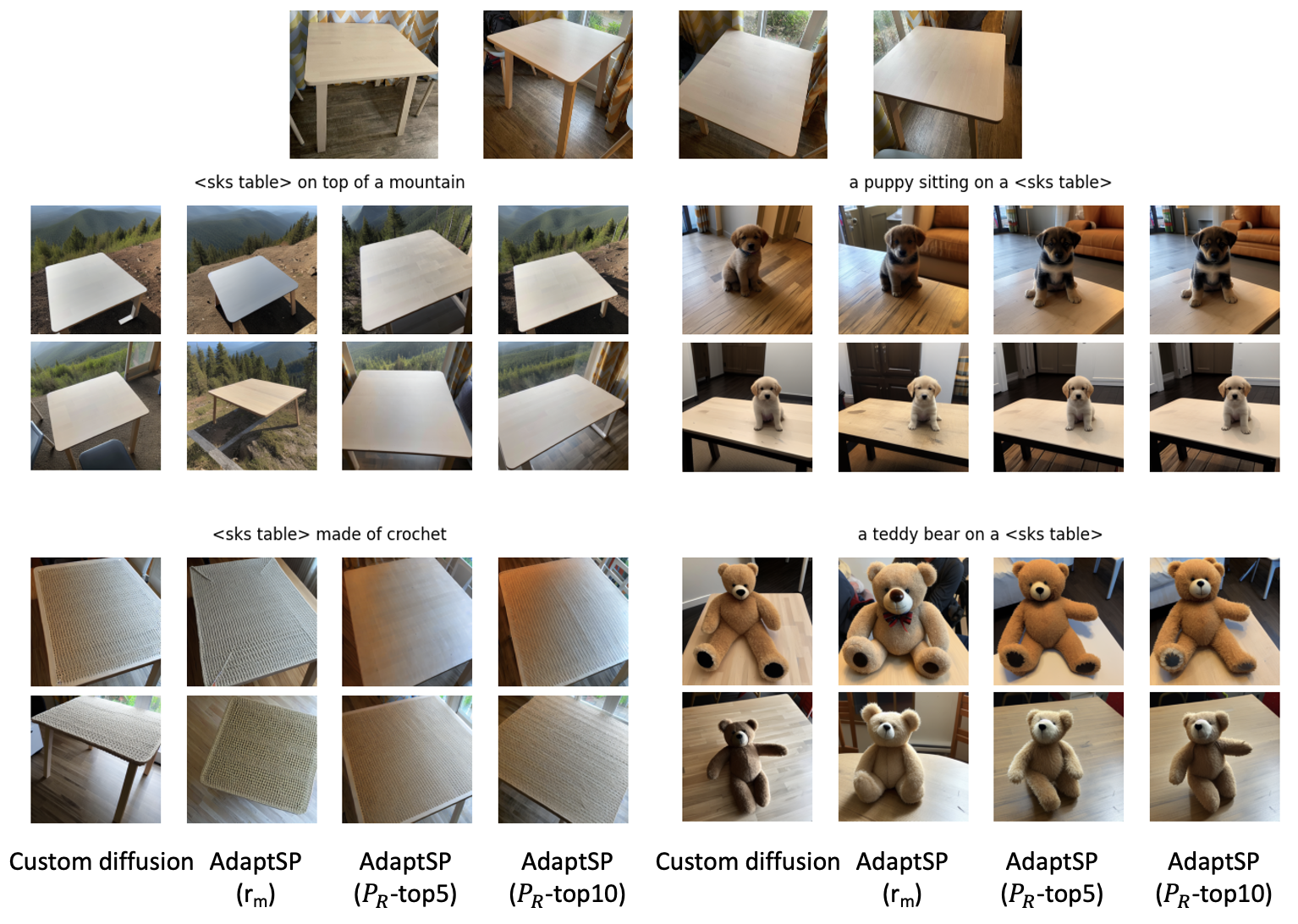}
    \caption{\textbf{Custom Diffusion on CC101 (concept table):} Qualitative comparison of Custom Diffusion and AdaptSP variants. The first row shows the reference images. The remaining rows show generations from different prompts and random seeds.}
    \label{fig:cs_cc101_table}
\end{figure}

In this section, we visualize and compare images generated by different methods using the same random seed with pre-trained Stable Diffusion v1.5. Figures~\ref{fig:db_celebA_342}, \ref{fig:db_celebA_908}, and \ref{fig:db_celebA_181} show DreamBooth results on three CelebA concepts, while Figures~\ref{fig:db_cc101_cat}, \ref{fig:db_cc101_teddy}, and \ref{fig:db_cc101_table} show DreamBooth results on three CC101 concepts. Figures~\ref{fig:cs_celebA_342}, \ref{fig:cs_celebA_908}, and \ref{fig:cs_celebA_181} present Custom Diffusion results on the same CelebA concepts, and Figures~\ref{fig:cs_cc101_cat}, \ref{fig:cs_cc101_teddy}, and \ref{fig:cs_cc101_table} present results on the three CC101 concepts. Since Class Diffusion shows similar trends to Custom Diffusion, we omit those examples here; additional qualitative results are provided in the supplementary material.

Overall, AdaptSP($r_m$) represents the personalized concept $V^*$ using a single residual vector while preserving its semantic appearance, supporting our first observation in Section~\ref{}. Across examples, AdaptSP substantially improves prompt fidelity and, in some cases, also enhances subject fidelity.

\section{Impact of subspace construction approach.}

To construct a meaningful semantic subspace that ensures the visual concept of interest ($V^*$ or its anchor concept $c$) is captured, the set of prompts must be chosen carefully. 
The guiding principle here is that the visual concept of interest acts as the central semantic thread common to all prompts, while the contexts are as diverse and varied as possible. 

This diversity is critical for isolating the core identity. For instance, if the prompt set frequently includes the phrase \texttt{a man wearing glasses}, the resulting principal components will incorrectly learn a composite concept of \texttt{a man with glasses} rather than the intended, context-free concept of \texttt{a man}. 
This is because the lack of contextual variance fails to distinguish the core identity from a recurring concept. 

Hence, we propose a set of prompts that pair the core concept ($V^*$ or $c$) with a wide range of contexts (e.g., actions, environments, artifacts, prompt length, etc.). By maximizing contextual variance, we ensure that the principal components of the embedding space capture the stable, high-variance signal of the core identity, isolating it from the lower-variance noise of individual contexts.
A set of such sample prompts are shown in Table \ref{tab:sample_prompts_subspace}.

\begin{table}[h!]
    \centering
    \caption{Sample prompts to construct the subspace. The data and prompts can be found in the supplementary material.}
    \label{tab:sample_prompts_subspace}
    \begin{tabular}{l|p{8cm}}
        \toprule
        \textbf{Set} & \textbf{Sample Sentences} \\
        \midrule
        \textbf{CelebA} & `A high-resolution photo of a \textcolor{red}{$V^*$}' \\
         & `A close-up portrait of a \textcolor{red}{$V^*$} smiling' \\
         & `A photo of a \textcolor{red}{$V^*$} looking tired but content after a long day' \\
         & `A photo of a \textcolor{red}{$V^*$} as a scientist in a laboratory, examining a petri dish' \\
         & `A watercolor painting of a \textcolor{red}{$V^*$}' \\
         & `A claymation figure of a \textcolor{red}{$V^*$}' \\
         & `A portrait of a \textcolor{red}{$V^*$} made from pressed flowers and leaves' \\
         & `A glowing, ethereal line art illustration of a \textcolor{red}{$V^*$}' \\
         & `A thermal image of a \textcolor{red}{$V^*$}' \\
        \midrule
        \textbf{CustomConcept101} & `A photo of a \textcolor{red}{$V^*$}' \\
         & `A rendering of a \textcolor{red}{$V^*$}' \\
         & `A cropped photo of a \textcolor{red}{$V^*$}' \\
         & `A portrait of a \textcolor{red}{$V^*$}' \\
         & `A close-up shot of a \textcolor{red}{$V^*$}' \\
         & `\textcolor{red}{$V^*$} next to a rustic wooden cabin' \\
         & `\textcolor{red}{$V^*$} covered in vines and moss' \\
         & `\textcolor{red}{$V^*$} in front of a glowing full moon' \\
         & `A cozy cabin with \textcolor{red}{$V^*$} sitting by the door' \\
        \bottomrule
    \end{tabular}
    \vspace{-2em}
\end{table}

% \subsection{Anchor choice}

% Compare $\tau_{\phi}(\lfloor p_i, c \lfloor)$ vs $\tau_{\phi_0}(\lfloor p_i, c \lfloor)$

% \subsection{}

% Idea: measure similarity between the secondary entity crop (how to automatically crop it?) and the reference set (sks man). If it’s too high, leakage is happening.

% proof for claim: Cross-class prompts: “kid”, “woman”, “old man”, etc. Track how much the output shifts toward the subject identity.

% Diversity metrics across multiple seeds and contexts: intra-set LPIPS, or DINO feature variance, or pairwise distance.

%%%%%%%%%%%%%%%%%%%%%%%%%%%%%%%%%%%%%%%%%%%%%%%%%%%%%%%%%%%%

\clearpage
\section*{NeurIPS Paper Checklist}

\begin{enumerate}

\item {\bf Claims}
    \item[] Question: Do the main claims made in the abstract and introduction accurately reflect the paper's contributions and scope?
    \item[] Answer: \answerYes{} % Replace by \answerYes{}, \answerNo{}, or \answerNA{}.
    \item[] Justification: Yes, the abstract and introduction accurately reflect our contributions and
scope.

\item {\bf Limitations}
    \item[] Question: Does the paper discuss the limitations of the work performed by the authors?
    \item[] Answer: \answerYes{} % Replace by \answerYes{}, \answerNo{}, or \answerNA{}.
    \item[] Justification: Yes, we include the discussion of limitations in the conclusion.
    \item[] Guidelines:
    \begin{itemize}
        \item The answer \answerNA{} means that the paper has no limitation while the answer \answerNo{} means that the paper has limitations, but those are not discussed in the paper. 
        \item The authors are encouraged to create a separate ``Limitations'' section in their paper.
        \item The paper should point out any strong assumptions and how robust the results are to violations of these assumptions (e.g., independence assumptions, noiseless settings, model well-specification, asymptotic approximations only holding locally). The authors should reflect on how these assumptions might be violated in practice and what the implications would be.
        \item The authors should reflect on the scope of the claims made, e.g., if the approach was only tested on a few datasets or with a few runs. In general, empirical results often depend on implicit assumptions, which should be articulated.
        \item The authors should reflect on the factors that influence the performance of the approach. For example, a facial recognition algorithm may perform poorly when image resolution is low or when images are taken in low lighting. Or a speech-to-text system might not be used reliably to provide closed captions for online lectures because it fails to handle technical jargon.
        \item The authors should discuss the computational efficiency of the proposed algorithms and how they scale with dataset size.
        \item If applicable, the authors should discuss possible limitations of their approach to address problems of privacy and fairness.
        \item While the authors might fear that complete honesty about limitations might be used by reviewers as grounds for rejection, a worse outcome might be that reviewers discover limitations that aren't acknowledged in the paper. The authors should use their best judgment and recognize that individual actions in favor of transparency play an important role in developing norms that preserve the integrity of the community. Reviewers will be specifically instructed not to penalize honesty concerning limitations.
    \end{itemize}

\item {\bf Theory assumptions and proofs}
    \item[] Question: For each theoretical result, does the paper provide the full set of assumptions and a complete (and correct) proof?
    \item[] Answer: \answerNA{} % Replace by \answerYes{}, \answerNo{}, or \answerNA{}.
    \item[] Justification: There is no theory in our method

    \item {\bf Experimental result reproducibility}
    \item[] Question: Does the paper fully disclose all the information needed to reproduce the main experimental results of the paper to the extent that it affects the main claims and/or conclusions of the paper (regardless of whether the code and data are provided or not)?
    \item[] Answer: \answerYes{}{} % Replace by \answerYes{}, \answerNo{}, or \answerNA{}.
    \item[] Justification: We provide the details of experiments in both main paper and appendix

\item {\bf Open access to data and code}
    \item[] Question: Does the paper provide open access to the data and code, with sufficient instructions to faithfully reproduce the main experimental results, as described in supplemental material?
    \item[] Answer: \answerYes{} % Replace by \answerYes{}, \answerNo{}, or \answerNA{}.
    \item[] Justification: We provide code in supplementary material with instructions.

\item {\bf Experimental setting/details}
    \item[] Question: Does the paper specify all the training and test details (e.g., data splits, hyperparameters, how they were chosen, type of optimizer) necessary to understand the results?
    \item[] Answer: \answerYes{} % Replace by \answerYes{}, \answerNo{}, or \answerNA{}.
    \item[] Justification: We specify the main experimental settings in the main paper and provide full details in Appendix \ref{sec:implemetation}.

\item {\bf Experiment statistical significance}
    \item[] Question: Does the paper report error bars suitably and correctly defined or other appropriate information about the statistical significance of the experiments?
    \item[] Answer: \answerNo{} % Replace by \answerYes{}, \answerNo{}, or \answerNA{}.
    \item[] Justification: We do not include error bars or statistical significance tests because all images are generated multiple times with different random seeds already, and we provide the average result, the same as the standard setting

\item {\bf Experiments compute resources}
    \item[] Question: For each experiment, does the paper provide sufficient information on the computer resources (type of compute workers, memory, time of execution) needed to reproduce the experiments?
    \item[] Answer: \answerNo{} % Replace by \answerYes{}, \answerNo{}, or \answerNA{}.
    \item[] Justification: We do not provide this information because the method is for a test-time optimization. However, we mentioned the inference time in the paper.
    
\item {\bf Code of ethics}
    \item[] Question: Does the research conducted in the paper conform, in every respect, with the NeurIPS Code of Ethics \url{https://neurips.cc/public/EthicsGuidelines}?
    \item[] Answer: \answerYes{} % Replace by \answerYes{}, \answerNo{}, or \answerNA{}.
    \item[] Justification: We respect the NeurIPS Code of Ethics.

\item {\bf Broader impacts}
    \item[] Question: Does the paper discuss both potential positive societal impacts and negative societal impacts of the work performed?
    \item[] Answer: \answerYes{} % Replace by \answerYes{}, \answerNo{}, or \answerNA{}.
    \item[] Justification: We provide the discussion of societal impacts in Appendix \ref{app:impact_statements}.
    
\item {\bf Safeguards}
    \item[] Question: Does the paper describe safeguards that have been put in place for the responsible release of data or models that have a high risk for misuse (e.g., pre-trained language models, image generators, or scraped datasets)?
    \item[] Answer: \answerNA{} % Replace by \answerYes{}, \answerNo{}, or \answerNA{}.
    \item[] Justification: The paper poses no such risks.

\item {\bf Licenses for existing assets}
    \item[] Question: Are the creators or original owners of assets (e.g., code, data, models), used in the paper, properly credited and are the license and terms of use explicitly mentioned and properly respected?
    \item[] Answer: \answerYes{} % Replace by \answerYes{}, \answerNo{}, or \answerNA{}.
    \item[] Justification: All the models and datasets from existing papers are properly credited, and their licenses are properly respected.

\item {\bf New assets}
    \item[] Question: Are new assets introduced in the paper well documented and is the documentation provided alongside the assets?
    \item[] Answer: \answerNA{}{} % Replace by \answerYes{}, \answerNo{}, or \answerNA{}.
    \item[] Justification: The paper does not release new assets.

\item {\bf Crowdsourcing and research with human subjects}
    \item[] Question: For crowdsourcing experiments and research with human subjects, does the paper include the full text of instructions given to participants and screenshots, if applicable, as well as details about compensation (if any)? 
    \item[] Answer: \answerNA{} % Replace by \answerYes{}, \answerNo{}, or \answerNA{}.
    \item[] Justification: Only authors join the evaluation process

\item {\bf Institutional review board (IRB) approvals or equivalent for research with human subjects}
    \item[] Question: Does the paper describe potential risks incurred by study participants, whether such risks were disclosed to the subjects, and whether Institutional Review Board (IRB) approvals (or an equivalent approval/review based on the requirements of your country or institution) were obtained?
    \item[] Answer: \answerNA{} % Replace by \answerYes{}, \answerNo{}, or \answerNA{}.
    \item[] Justification: Only authors join the evaluation process, so no potential risk

\item {\bf Declaration of LLM usage}
    \item[] Question: Does the paper describe the usage of LLMs if it is an important, original, or non-standard component of the core methods in this research? Note that if the LLM is used only for writing, editing, or formatting purposes and does \emph{not} impact the core methodology, scientific rigor, or originality of the research, declaration is not required.
    %this research? 
    \item[] Answer: \answerYes{} % Replace by \answerYes{}, \answerNo{}, or \answerNA{}.
    \item[] Justification: The core method development in this research does not involve LLMs as any important, original, or non-standard components. However we use LLM to generate data.

\end{enumerate}

\end{document}